\documentclass{article}
\pdfoutput=1

\usepackage{amsmath,amsfonts,bm}

\def\eqref#1{equation~\ref{#1}}

\def\1{\bm{1}}

\DeclareMathAlphabet{\mathsfit}{\encodingdefault}{\sfdefault}{m}{sl}
\SetMathAlphabet{\mathsfit}{bold}{\encodingdefault}{\sfdefault}{bx}{n}

\DeclareMathOperator*{\argmin}{arg\,min}

\usepackage{amsmath}
\usepackage{color}
\usepackage{pgfplots, pgfplotstable}
\usepackage[colorinlistoftodos]{todonotes} 

\newcommand{\loss}{MC-GTA}
\newcommand{\model}{MC-GTA}
\newcommand{\modelfull}{Goodness-of-fit Tests with Autocorrelations}

\newcommand{\maxzero}[1]{\lfloor #1 \rfloor_+}

\newcommand{\pboxtwo}[2]{\pbox{2cm}{#1 \newline #2}}

\newcommand{\modelobjfun}{\mathcal{L}^{\text{\model}}}
\newcommand{\mcobjfun}{\mathcal{L}^{\text{mc}}}
\newcommand{\mcmbobjfun}{\mathcal{L}^{\text{mcm}}}
\newcommand{\metricpen}{r}
\newcommand{\empE}{\widehat{\mathbb{E}}}

\newcommand{\metricspa}{M}
\newcommand{\metricdis}{d_c}

\newcommand{\clus}{C}
\newcommand{\clusassign}{\mathcal{C}}
\newcommand{\cluspred}{\hat{\mathcal{C}}_K}
\newcommand{\metricontcluspred}{\hat{\mathcal{C}}_K^{\text{mc}}}

\newcommand{\dataset}{\mathcal{D}}

\newcommand{\obs}{X}
\newcommand{\obsnum}{N}
\newcommand{\feavec}{\mathbf{f}}
\newcommand{\feadim}{\mathrm{d}_{F}}

\newcommand{\posvec}{\mathbf{p}}
\newcommand{\posdim}{\mathrm{d}_{M}}

\newcommand{\mrf}{\mathcal{M}}
\newcommand{\mrfparam}{\mathbf{\theta}}

\newcommand{\feavdistance}{d_f}
\newcommand{\mbdistance}{d_m}
\newcommand{\mrfdistance}{W_2}

\newcommand{\rebut}[1]{\textcolor{black}{{#1}}}

\usepackage{hyperref}
\usepackage{url}
\usepackage[utf8]{inputenc} 
\usepackage[T1]{fontenc}    
\usepackage{hyperref}       
\usepackage{url}            
\usepackage{booktabs}       
\usepackage{amsfonts}       
\usepackage{nicefrac}       
\usepackage{microtype}      
\usepackage{xcolor}         
\usepackage{graphicx}
\usepackage{mathtools}
\usepackage{amsmath}
\usepackage[linesnumbered,ruled,algo2e]{algorithm2e}
\usepackage{enumitem}
\usepackage{float}
\usepackage[caption = true]{subfig}
\usepackage{bbm}
\usepackage{xcolor}
\usepackage{pbox}

\usepackage{microtype}
\usepackage{graphicx}
\usepackage{booktabs} 
\usepackage{multirow}

\usepackage{hyperref}

\usepackage[accepted]{icml2024}

\usepackage{amsmath}
\usepackage{amssymb}
\usepackage{mathtools}
\usepackage{amsthm}

\usepackage[capitalize,noabbrev]{cleveref}

\theoremstyle{plain}

\theoremstyle{definition}

\theoremstyle{remark}

\usepackage{todonotes}

\icmltitlerunning{\model: Metric-Constrained Model-Based  Clustering using \modelfull}

\begin{document}

\twocolumn[
\icmltitle{\model: Metric-Constrained Model-Based  Clustering using \modelfull}



\icmlsetsymbol{equal}{*}

\begin{icmlauthorlist}
\icmlauthor{Zhangyu Wang}{ucsb}
\icmlauthor{Gengchen Mai}{uga,utaustin}
\icmlauthor{Krzysztof Janowicz}{uwien,ucsb}
\icmlauthor{Ni Lao}{google}
\end{icmlauthorlist}

\icmlaffiliation{ucsb}{Department of Geography, University of California Santa Barbara, CA, USA}
\icmlaffiliation{uga}{Department of Geography, University of Georgia, GA, USA}
\icmlaffiliation{utaustin}{SEAI Lab, Department of Geography and the Environment, University of Texas at Austin, TX, USA}
\icmlaffiliation{uwien}{Faculty of Geosciences, Geography and Astronomy, University of Vienna, Vienna, Austria}
\icmlaffiliation{google}{Google, Mountain View, CA, USA}

\icmlcorrespondingauthor{Zhangyu Wang}{zhangyuwang@ucsb.edu}
\icmlcorrespondingauthor{Gengchen Mai}{gengchen.mai@austin.utexas.edu}
\icmlcorrespondingauthor{Krzysztof Janowicz}{krzysztof.janowicz@univie.ac.at}
\icmlcorrespondingauthor{Ni Lao}{noon99@gmail.com}

\icmlkeywords{Machine Learning, ICML}

\vskip 0.3in
]



\printAffiliationsAndNotice{}  

\begin{abstract}

A wide range of (multivariate) temporal (1D) and spatial (2D) data analysis tasks, such as grouping vehicle sensor trajectories, can be formulated as 
clustering with given metric constraints. 
\rebut{Existing metric-constrained clustering algorithms 
overlook the rich correlation between feature similarity and metric distance, i.e., metric autocorrelation. 
The model-based variations of these clustering algorithms (e.g. TICC and STICC) achieve SOTA performance,
yet suffer from computational instability and complexity by using a metric-constrained
Expectation-Maximization procedure.}
\rebut{In order to address these two problems, we propose a novel clustering algorithm, \model{} (\textbf{M}odel-based \textbf{C}lustering via \textbf{G}oodness-of-fit \textbf{T}ests with 
\textbf{A}utocorrelations). 
Its objective is only composed of pairwise weighted sums of feature similarity terms (square Wasserstein-2 distance) and metric autocorrelation terms (a novel multivariate generalization of classic semivariogram). 
We 
show that 
\model{} is effectively minimizing the total hinge loss for intra-cluster observation pairs not passing goodness-of-fit tests, i.e., statistically not originating from the same distribution.}
Experiments on 1D/2D synthetic and real-world datasets demonstrate that \model~ successfully incorporates metric autocorrelation.
It outperforms strong baselines by large margins (up to 14.3\% in ARI and 32.1\%  in NMI) 
with faster and stabler optimization \rebut{(>10x speedup)}.

\end{abstract}

\section{Introduction}   \label{sec:intro}

\begin{figure*}[t!]
\centering
\includegraphics[width = 0.99\textwidth]{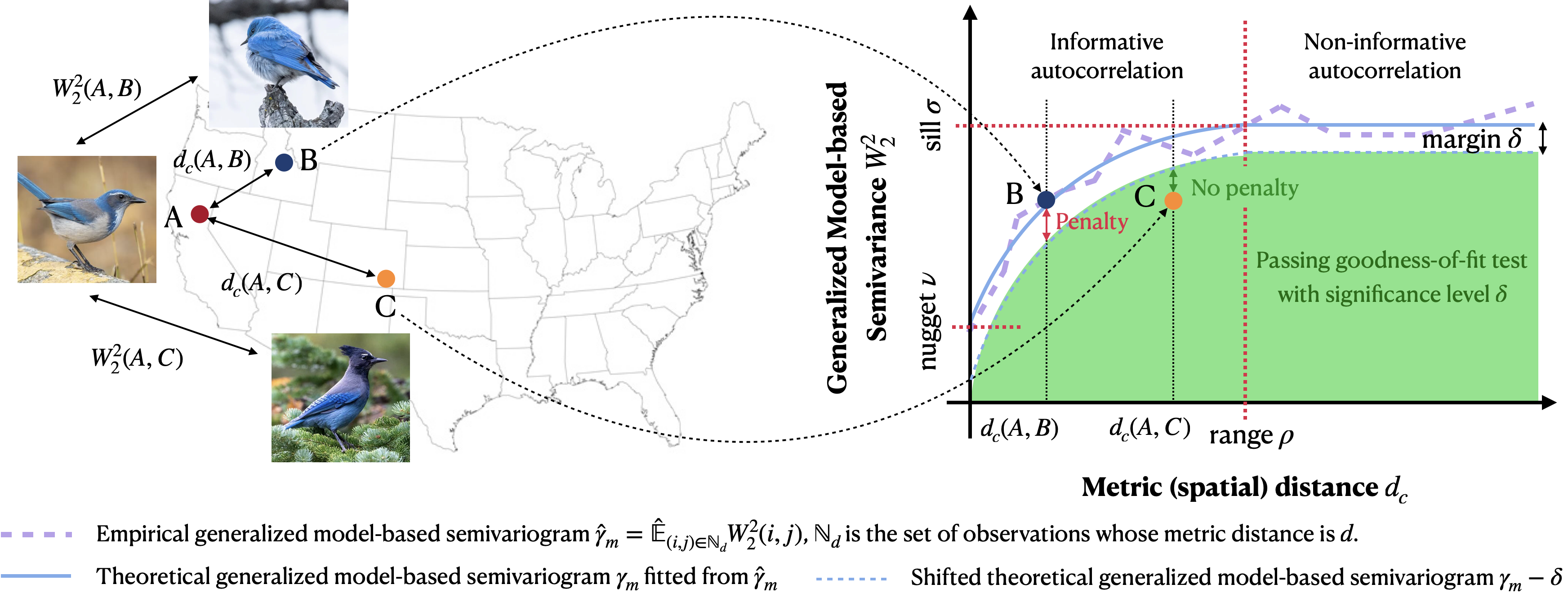}
\vspace{-0.4cm}
\caption{Motivation of \model{} using iNaturalist-2018 dataset as an example. 
We wish to cluster wild animal photos based on both their image similarity and spatial adjacency. For any pair of observations, we obtain their metric distance $d_c$ and generalized model-based semivariance $\mrfdistance^2$ (square Wasserstein-2 distance), which quantifies feature similarity via underlying models. In the presence of metric autocorrelation, the expected generalized model-based semivariance is in theory an increasing function of $d_c$ within range $\rho$ and levels off beyond $\rho$, namely \textit{theoretical generalized model-based semivariogram} $\gamma_m$. We fit $\gamma_m$ from the empirical generalized model-based semivariogram $\hat{\gamma}_m$. \model{} penalizes observation pairs whose $\mrfdistance^2$ is close to or exceeding $\gamma_m$ via a hinge loss with margin $\delta$. An observation pair having no hinge loss penalty equals passing a goodness-of-fit test with significance level $\delta$. 
}
\label{fig:intro-eyecatch}
\vspace{-0.2cm}
\end{figure*}

Clustering is one of the most fundamental problems in unsupervised learning, which deals with the data partitioning when ground-truth labels are unknown \citep{xu2015comprehensive}. Most existing clustering algorithms only consider the similarity among observations in the feature (attribute) space. 
However, in real-world applications, additional \textit{metric constraints} (e.g., temporal continuity and geospatial proximity) often matter, especially in temporal and spatial data mining~\citep{BIRANT2007208,hu2015extracting,mai2018adcn,BELHADI2020103857}. 
In other words, observations have both features and positions in a metric space. Hence, observations that are put in the same cluster should be similar in terms of features \textbf{and} their positions in the metric space (e.g., timestamps or geographic locations) should also satisfy some constraints.
Metric constraints can be generalized to even higher dimensions as long as a meaningful distance measure is defined. 
This kind of problems is commonly known as \textit{metric-constrained clustering} \citep{veldt2019metric}.
%
It is \textbf{not} a trivial task to design a larger composite space from these two spaces 
because they may have completely different metrics. For example, concatenating word embedding with geo-coordinates makes learning good similarity functions difficult because the former uses cosine distance while the latter uses Euclidean/geodesic distance, whose values can not be directly compared.


The current state-of-the-art metric-constrained clustering algorithms, namely TICC~\citep{hallac2017ticc} (for temporal clustering) and STICC~\citep{kang2022sticc} (for spatial clustering), consider both spaces by combining model-based clustering~\citep{gormley2022} 
with a soft metric penalty
%
\(
\argmin_{\Theta, \clusassign}  \sum_{k=1}^K \Big[ \big\Arrowvert\lambda \circ \mrfparam_{C_k}\big\Arrowvert_1 + 
\sum_{X_i \in C_k} \big( -ll(X_i, \mrfparam_{C_k}) + \beta \mathbbm{1}\{\Tilde{X_{i}} \notin C_k\}\big)\Big]
\).
Here $K$ is the total number of clusters. $X_i$ is one observation we need to assign to a cluster, and $\Tilde{X_{i}}$ is the nearest neighbor of $X_i$ in the metric space. $\mrfparam_{C_k}$ are the estimated model parameters for cluster $k$, $C_k$ is the set of observations of cluster $k$, $-ll(X_i,\mrfparam_{C_k})$ is the negative log-likelihood of observation $X_i$  belonging to cluster $C_k$ given model parameter $\mrfparam_{C_k}$, and $\beta \mathbbm{1}\{\Tilde{X_{i}} \notin C_k\}$ is the soft metric penalty. $\lambda$ is an L-1 normalization \rebut{hyper-parameter to prevent overfitting.}
The advantage of this strategy is three-fold. Firstly, similarity computed based on underlying models is by nature more robust to noise and outliers than that based on raw feature vectors ~\citep{wang2016functional}. 
Secondly, estimated underlying models provide better interpretability~\citep{hallac2017ticc}. 
Finally, the magnitude of penalty can be tuned to adjust the emphasis on metric constraints, preferable to methods that enforce metric constraints as hard rules, such as ST-DBSCAN~\citep{BIRANT2007208}, MDST-DBSCAN~\citep{ijgi10060391}, and 
semi-supervised 
algorithms that discretize the metric constraints into graphs~\cite{10.5555/645530.655669, basu2004active, lu2007semi, DBLP:journals/corr/abs-1907-10410, boecking2022constrained}.
However, these approaches also have major drawbacks.

The most critical weakness of 
all existing clustering algorithms is that they ignore the effects of metric autocorrelation, e.g., temporal/spatial autocorrelation~\citep{GoodchildMichaelF1987Sa,anselin1988spatial,fortin2002spatial,gubner_2006}, when applying the metric constraints. Metric autocorrelation effectively asserts that within a cluster, feature vectors observed at metrically distant positions \rebut{naturally} 
have higher empirical variance than metrically adjacent ones. Figure \ref{fig:intro-eyecatch} shows how considering metric autocorrelation may even reverse the clustering results: \rebut{suppose observation $B$ and observation $C$ are equally similar to observation $A$ feature-wise, but $C$ is farther away from $A$ than $B$. Without metric autocorrelation, $B$ should be preferred to be clustered with $A$, since it has smaller metric distance; but if we do consider metric autocorrelation, 
$C$ should be preferred instead, because autocorrelation implies that $C$ would have been more similar to $A$ if it were in $B$'s metric position.}
Using log-likelihood as a clustering objective makes it hard to integrate metric constrains in a generative process.
Although log-likelihood alone as loss function fits well with model selection theories such as Akaike Information Criterion (AIC), the sum of log-likelihood and weighted distance penalties 
lacks statistical meaning. 
Moreover, the presence of the penalty term breaks the convergence guarantee of EM iterations. 
Empirically, TICC/STICC is highly non-convex and difficult to optimize~\citep{hallac2017ticc, kang2022sticc}.
Other issues include high computational complexity, sensitivity to initial conditions, and expensive hyperparameter tuning.
Please refer to Section \ref{sec:rel-works} and Appendix \ref{sec:runtime} for detailed discussions. These problems, however, can be avoided by getting rid of log-likelihood and EM iterations. 
One strategy is to 
perform clustering only according to pairwise similarity measurements, similar to DBSCAN.

In this paper, inspired by the analysis above, we propose a novel model-based clustering method named \model~ 
(Model-based Clustering via \modelfull)
that expilicitly accounts for the metric autocorrelation by designing a Wasserstein-2 distance-based multivariate generalization of the spatial semivariogram, which is widely used in geostatistics~\citep{isaaks1989applied}. 
\model~ first fits an underlying model (Gaussian Markov Random Field) for each observation using its neighbors. Then we compute the generalized model-based semivariance (i.e., square Wasserstein-2 distance) and metric distance for all observation pairs. Next, we fit the theoretical generalized model-based semivariogram and form our clustering objective as a total hinge loss based on the difference between the empirical semivariance and the theoretical semivariogram. This injects metric constraints into clustering. Finally, we develop an algorithm that minimizes the loss. We prove that our objective can be theoretically interpreted in terms of goodness-of-fit tests.
We use extensive experiments to demonstrate that \model~ can mitigate the computational complexity and convergence instability problems of TICC/STICC, achieving significantly better clustering quality.



To summarize, the \textbf{major contributions} of this paper are:
\setlist[itemize]{leftmargin=5.5mm}
\begin{itemize}[topsep=0pt,parsep=0pt,partopsep=0pt] 
    \item We propose a Wasserstein-2 distance-based generalization of semivariogram that explicitly accounts for multivariate metric autocorrelation. 
    \item We propose a novel model-based clustering objective based on goodness-of-fit tests. It simultaneously enables incorporating metric autocorrelation information and improves computational stability/efficiency. 
    We believe that clustering based on statistic tests is a promising direction for future research developments.
    \item We compare our method with existing works comprehensively on various 1D and 2D synthetic and real-world datasets. 
    We demonstrate that our method outperforms the baselines in clustering quality, computational stability, and computational efficiency.
\end{itemize}

\section{Related Works}\label{sec:rel-works}




\paragraph{Spatial and Temporal Clustering.}
Clustering temporal subsequences and spatial subregions is a well-studied sub-field of clustering (Appendix \ref{sec:app_general_cluster_obj}). Some works treat temporal/spatial information as indices, such as dynamic time warping~\citep{10.1145/2783258.2783286,10.5555/1287369.1287405,10.1145/347090.347153,10.1145/2339530.2339576}, time point clustering~\citep{10.1145/640075.640091,review-of-subsequence} and geo-tagged images~\citep{rs10040654}, and some works cluster the spatio-temporal trajectories directly~\citep{BELHADI2020103857,kisilevich2010spatio}. We are mostly interested in the first case, i.e., clustering temporally/geospatially referenced observations. However, these methods generally perform clustering based on feature similarity, which can be problematic or even unreliable~\citep{1250910}, because it only considers the structure of features, ignoring that the observations are also distributed over time and space.

To address this problem, two main strategies are explored in previous works. The first strategy is to enforce metric constraints as hard rules. For example, in ST-DBSCAN~\citep{BIRANT2007208} and MDST-DBSCAN~\citep{ijgi10060391}, only temporally dense observations are considered candidates for core observations. The second strategy is to add a soft metric penalty to the clustering optimization objective. TICC~\citep{hallac2017ticc} is the first work to introduce Markov Random Fields to model temporal dependency structures of subsequences together with a soft temporal penalty. STICC~\citep{kang2022sticc}, following this work, modified the algorithm to suit 2-dimensional spatial subregion clustering. They are both model-based clustering algorithms, like ARMA~\citep{XIONG20041675}, GMM~\citep{mclust} and HMMs~\citep{10.5555/2998981.2999073}. TICC/STICC achieves state-of-the-art performance in temporal/spatial clustering tasks.

\section{Problem Formulation}\label{sec:problem-setup}

\subsection{
Metric-Constrained Clustering}\label{sec:metric-constraint}

Given a dataset $\dataset$ of $\obsnum$ observations $\{\obs_i\}_{i=1}^\obsnum$ (e.g., points of interest in an urban area, sensor measurements at different time points, etc), we need to assign 
each $\obs_i$ to a set $\clus_k$, i.e. cluster $k$. The set of all clusters $\clusassign = \{\clus_k\}_{k=1}^K$ is a \textit{cluster assignment} or a \textit{clustering}. $K$ is called the number of clusters, either predefined or inferred from data.

Each observation $\obs_i = (\feavec_i, \posvec_i)$ is a tuple of two vectors: $\feavec_i$ is a $\feadim$-dimensional feature vector (e.g., attributes of a POI) in a feature space $F$, while $\posvec_i$ is a $\posdim$-dimensional position vector (e.g., geo-coordinates of this POI) in a metric space $(\metricspa, \metricdis)$ (e.g., Earth surface with geodesic distance), where $\metricdis$ is a predefined metric. 
With a dissimilarity measurement $\feavdistance(\cdot, \cdot)$ in the feature space, e.g., cosine distance, a classic clustering problem without metric constraints $\cluspred 
=\argmin_{\clusassign} \mathcal{L(\clusassign)}$ is to minimize the loss:
\begin{align}
\begin{split}
\mathcal{L(\clusassign)} =&
 \sum_{\{C_k \in \clusassign\}}
 \sum_{\{i,j\in C_k\}} \feavdistance(\feavec_i,\feavec_j) \\
& - \alpha \sum_{\{C_k, C_l \in \clusassign, k \neq l\}}\sum_{\{i\in C_k, j \in C_l\}}\feavdistance(\feavec_i,\feavec_j)
\end{split}
\end{align}
where the first term is the intra-cluster cohesion objective and the second is the inter-cluster separation objective. $\alpha$ is a hyperparameter balancing cohesion and separation.
\rebut{
Many applications
emphasize more on intra-cluster cohesion. Following TICC/STICC \citep{hallac2017ticc,kang2022sticc}, we set $\alpha=0$ in this study.} 

A \textit{metric constraint} is an additional loss $\mcobjfun$ that assigns penalty based on metric distance and feature similarity. A \textit{metric-constrained clustering} problem is to find an optimal cluster assignment that minimizes a  multi-objective $\metricontcluspred =\argmin_{\clusassign} \Big[\mathcal{L}(\clusassign) + \beta \mcobjfun(\clusassign) \Big] $ where $\beta$ is a hyperparameter that determines how soft the constraints are, and
\begin{equation}
\begin{split}
\mcobjfun(\clusassign) =\sum_{\{C_k \in \clusassign\}}
\sum_{\{i,j\in C_k\}} r(\feavdistance(\feavec_i, \feavec_j), \metricdis(\posvec_i,\posvec_j)).
\end{split}
\label{eq:metric_cont}
\end{equation}
$r$ is a function of the metric distance and the feature dissimilarity, called the \textit{metric penalty function}, designed to properly enforce the metric constraints.  For example, in ST-DBSCAN~\citep{BIRANT2007208}, temporal continuity is the metric constraint. Conceptually it corresponds to $r(\feavdistance(\feavec_i, \feavec_j), \metricdis(\posvec_i,\posvec_j)) = \mathbbm{1}\{\metricdis(\posvec_i,\posvec_j) > \epsilon_t\}$ with $\epsilon_t$ being the preset radius of temporal neighborhood, and $\beta=\infty$. It effectively means that cluster assignments with temporal discontinuity are hard eliminated.

\subsection{Metric-Constrained Model-Based Clustering}\label{sec:metric-constrained-model-based}

Metric-Constrained Model-based (MCM)  clustering is a special case of metric-constrained clustering, which
views the feature vector $\feavec_i$ of observation $\obs_i$ as a random sample drawn from a parametric distribution $\mrf(\mrfparam_i)$. 
We say $\mrf(\mrfparam_i)$ is the underlying model of $\obs_i$ and $\mrfparam_i$ is a specification of the parameters. The family of distribution (e.g. Gaussian) and exact parameterization (e.g. mean and covariance matrix) of $\mrf(\mrfparam_i)$ is chosen a priori based on domain knowledge and computational considerations. In MCM clustering, the feature dissimilarity measure $\feavdistance(\feavec_i, \feavec_j)$ is replaced with $\mbdistance(i,j)  = \mbdistance(\feavec_i, \feavec_j, \mrfparam_i, \mrfparam_j; \mrf)$, named as \textit{model-based dissimilarity}, e.g. negative log-likelihood. 

In summary, MCM clustering can be formulated as minimizing the 
MCM loss\footnote{
For the rest of the paper, 
we abbreviate the  notations by omitting the arguments and only keeping the indices. For example, $\mbdistance(\feavec_i, \feavec_j, \mrfparam_i, \mrfparam_j; \mrf)$ 
is written in short as $\mbdistance(i,j)$, $\metricdis(\posvec_i, \posvec_j)$ as $\metricdis(i, j)$, $r(\mbdistance(i, j), \metricdis(i, j))$ as $r(i,j)$, respectively.
}:
\begin{align}\label{eq:mcmb}
\begin{split}
\mcmbobjfun(\clusassign) = 
& \sum_{\{C_k \in \clusassign\}}
 \sum_{\{i,j\in C_k\}} [\mbdistance(i,j) + \beta r(i,j)] 
\end{split}
\end{align}
As TICC authors~\cite{hallac2017ticc}  argued, distance-based metrics have been shown to yield unreliable results in certain situations. While model-based approaches prevents overfitting and allows us to discover types of patterns that other approaches are unable to find. 
See Section~\ref{sec:abla-sim} for a detailed analysis of the tasks in this study.

\section{Preliminaries}\label{sec:preliminaries}
In this section, we introduce a few useful statistical tools for our proposed clustering algorithm.

\subsection{Classic Univariate Semivariogram}\label{sec:gsemivariogram}

In Section \ref{sec:intro}, we argued for the importance of metric autocorrelation in metric-constrained clustering. 
In order to incorporate it into the clustering process, we need to appropriately quantify it. While an abundance of statistics for autocorrelation tests are developed in classic temporal and spatial analysis, such as Durbin–Watson statistic~\citep{157690d6-0008-3f13-8b5b-c7bccf4730c5,3786e5c0-cb68-389d-af38-9300b4442a9c} and Moran's I~\citep{af54b142-01be-3f96-af47-1730365d8376}, the semivariogram~\citep{10.2113/gsecongeo.58.8.1246} fits our end best. This is because the theoretical semivariogram, denoted as $\gamma(\posvec_i,\posvec_j)$, is a function describing the degree of spatial dependence of a spatial random field or stochastic process, which is literally the fundamental assumption of model-based clustering. 

Given a dataset (which is a sample generated from the spatial stochastic process) of $N$ univariate observed variables $\{z_1,\cdots z_N\}$ together with their spatial positions $\{\posvec_1,\cdots \posvec_N\}$, there are $N^2$ pairs of variables $(z_i,z_j)$ and their corresponding pairs of spatial positions $(\posvec_i,\posvec_j)$. The empirical semivariogram is defined as 
\begin{equation}\label{eq:emp-variogram}
\begin{split}
\hat{\gamma}(h\pm \epsilon) := 
\dfrac{1}{2|N(h\pm\epsilon)|}
\sum_{\{(\posvec_i,\posvec_j)\in N(h\pm\epsilon)\}}|z_i-z_j|^2
\end{split}
\end{equation}
where $N(h\pm\epsilon) := \{(\posvec_i,\posvec_j)| h-\epsilon \leq \metricdis(\posvec_i,\posvec_j) \leq h+\epsilon\}$, a set of spatial positions, and $|N(h\pm\epsilon)|$ is the size of the set. This is essentially the half empirical variance of all pairs whose spatial distance falls into the same distance bin centered at $h$ of width $2\epsilon$. 

In a semivariogram, the x and y axes indicate the spatial distance and semivariance $\hat{\gamma}(h\pm \epsilon)$, respectively. 
In the beginning the semivariance rises as distance increases, which indicates spatial autocorrelation. Then it levels off, which indicates that now semivariance no longer provides useful information. The range $\rho$ is the distance beyond which spatial autocorrelation levels off. The sill $\sigma$ is the semivariance when spatial autocorrelation levels off. The nugget $\nu$ is the semivariance when distance is almost zero, which is considered an intrinsic variance of the stochastic process.

Semivariogram can also be applied to metric spaces other than 2D or 3D geospatial space, such as temporal space, spatio-temporal space, and even multi-dimensional, non-Euclidean spaces~\citep{amt-7-2631-2014}. However, they are designed to quantify the autocorrelation between univariate observations and 
can not be applied to multivariate cases as shown in Figure \ref{fig:intro-eyecatch}. This is because the concepts of range, sill and nugget are defined as turning/intercepting points of the function. If $\gamma$ is multivariate, the three core concepts are not well-defined. One of our novel contributions is to generalize the definition of semivariogram to multivariate observations. Figure \ref{fig:intro-eyecatch} illustrates this generalization.

\subsection{Wasserstein-2 Distance and Gaussian Markov Random Fields}\label{sec:model-similarity}

To generalize semivariograms to multivariate cases, our strategy is to replace the univariate distance in Equation \ref{eq:emp-variogram} with a model dissimilarity measurement.
There are various statistical metrics or quasi-metrics that can be used, such as divergence (such as KL-divergence), total variation, discrepancy, and Wasserstein-2 distance~\citep{gibbs2002choosing}. 
We wish to choose one that is compatible with the classic semivariogram's definition. Specifically, we need to show that having a small semivariance in terms of model dissimilarity guarantees having a small semivariance in terms of feature difference. The weakest possible condition that satisfies this \rebut{requirement} is weak convergence, also known as convergence in distribution. Intuitively it says if a model weakly converges to another model, the observations generated from them will become statistically indistinguishable, consequently having indistinguishable semivariance. Therefore, we need to find a statistical metric $\mrfdistance$ that metricizes weak convergence, i.e.,
    $(\mrfdistance(i, j) \rightarrow 0) \Rightarrow (\mathcal{F}_i \xrightarrow[\text{}]{\text{D}} \mathcal{F}_j)$.
Here $\mathcal{F}_i, \mathcal{F}_j$ are cumulative distribution functions parametrized by $\mrfparam_i, \mrfparam_j$ respectively, and $\xrightarrow[\text{}]{\text{D}}$ denotes convergence in distribution.
Among all such metrizations, L\'evy-Prokhorov metric and Wasserstein's distance (Earth Mover's Distance) are the two most important cases. By \cite{gibbs2002choosing}, L\'evy-Prokhorov metric is the tightest bound of the distance between two distributions, and the Wasserstein's distance is only looser up to a constant factor. Please refer to Appendix \ref{app:lpm} for the definition of both distance metrics.

Whereas the L\'evy-Prokhorov Metric is in general not computable, the square Wasserstein-2 distance between two Gaussian Markov Random Fields (GMRFs) has a beautiful closed-form:
\begin{equation}\label{eq:w2}
    W_2^2(\mrfparam_i, \mrfparam_j) = d_2^2(\mu_i, \mu_j) + Tr\big(\Sigma_i + \Sigma_j - 2A\big)
\end{equation}
Here $\mu_i, \mu_j$ are mean vectors, $\Sigma_i, \Sigma_j$ are covariance matrices, and $\mrfparam_i = (\mu_i, \Sigma_i)$. 
$A=(\Sigma_i^{1/2}\Sigma_j \Sigma_i^{1/2})^{1/2}$ and
$Tr(\cdot)$ is the trace of a matrix. $d_2^2$ is the square L2 norm. For the simplicity of notations, we use $W_2^2(i,j)$ in abbrevation of $W_2^2(\mrfparam_i, \mrfparam_j)$ throughout this paper.

The analysis above demonstrates that the combination of Wasserstein-2 distance and GMRFs 
is essentially the only choice we have that both satisfies our requirement of weak convergence and comes with practical computability.
Please refer to Appendix \ref{sec:gmrf} for more background about GMRF. In Section \ref{sec:methodology} we show that our clustering objective based on square Wasserstein-2 distance has  clear statistical meaning.

\section{Method}\label{sec:methodology}

\subsection{Generalized Model-based Semivariogram}\label{sec:general-semivariogram}

We propose a novel multivariate generalization of the classic semivariogram, called \textit{generalized model-based semivariogram}, to appropriately quantify the multivariate metric autocorrelation. Unlike existing work such as \citet{Abzalov2016}, which modifies the definitions of range, sill, and nugget analogously to simultaneous confidence intervals, we derive a natural generalization by replacing the variance $|z_i-z_j|^2$ in Equation \ref{eq:emp-variogram} with $\mbdistance := \mrfdistance^2$. As in model-based clustering, every observed feature vector $\feavec_i$ has an underlying model $\mrf(\mrfparam_i)$. Though the difference between the feature vectors is multivariate, the difference between the underlying models is univariate. 
We define the empirical generalized model-based semivariogram $\hat{\gamma}_m$ as
\begin{equation}\label{eq:generalized-emp-variogram}
\begin{split}
\hat{\gamma}_m(h\pm \epsilon) := \dfrac{1}{2|N(h\pm\epsilon)|}
\sum_{(\posvec_i,\posvec_j)\in N(h\pm\epsilon)} 
\mrfdistance^2(i,j)
\end{split}
\end{equation}
Following that, we can fit a theoretical generalized model-based semivariogram $\gamma_m$ on $\hat{\gamma}_m$ by using well-established, classic univariate semivariogram fitting  methods~\citep{MULLER199993}. $\rho$ is the range of the fitted theoretical semivariogram.\footnote{
For simplicity, in the rest of the paper, when mentioning semivariogram/semivariance, we always refer to the generalized model-based definitions unless otherwise specified.
}

The soundness of this generalized definition \rebut{is theoretically supported by goodness-of-fit tests (Section \ref{sec:mcmb-good-of-fit})}. It is also verified on real-world datasets. 
Figure \ref{fig:variogram} is the empirical generalized model-based semivariogram computed on a large geo-tagged image dataset iNaturalist-2018~\citep{cui2018large}. We use the top 16 PCA components of the pretrained image embedding as the feature vector, and the distance is the great circle distance between the geo-tags. For each image, we use its 15/20/30-nearest neighbors to estimate a GMRF as the underlying model. 
Then we compute the generalized semivariogram using Equation \ref{eq:generalized-emp-variogram}. We can see the empirical semivariogram conforms very well with the theory.


\subsection{Clustering Objective as Goodness-of-Fit Tests}\label{sec:mcmb-good-of-fit}

The conventional likelihood-based EM iterations of model-based clustering algorithms malfunction when metric constraints are involved.  
\rebut{To avoid this situation, we propose to formulate the loss function in terms of only pairwise computations between observations. 
Our solution mainly relies on goodness-of-fit tests (i.e., whether two samples come from a statistically identical distribution) as the pairwise computation. 
We punish the pairs that have large goodness-of-fit test statistics beyond a significance threshold with a hinge loss. 
This threshold is based on the average goodness-of-fit test statistic dependent on metric distance. }
\rebut{Specifically, the generic $\mcmbobjfun$ loss (Equation \ref{eq:mcmb}) can be realized with goodness-of-fit tests as follows: 
}
{ 
\begin{equation}\label{eq:mcmb-as-test}
\begin{split}
&\modelobjfun(\clusassign)= 
 \sum_{\{C_k \in \clusassign\}}
 \sum_{\{i,j\in C_k\}}\Bigl[ \\
 &\quad
 \maxzero{
 \mrfdistance^2(i,j) - (\empE_{i^{'},j^{'} \in \mathbb{N}} \mrfdistance^2(i',j') - \delta^{0}) } 
 \\
 &\quad
 + \beta \maxzero{
 \mrfdistance^2(i,j) - (\empE_{i^{'},j^{'}\in \mathbb{N}_{i,j}} \mrfdistance^2(i',j') - \delta) }
 \Bigr]
\end{split}
\end{equation}
}
\rebut{Here $\mathbb{N}$ is the set of all observation pairs, and  $\mathbb{N}_{i,j}=N(d_c(i,j)\pm\epsilon)$ are the 
pairs in $(i,j)$'s distance bin as defined in the generalized model-based semivariogram (Section~\ref{sec:general-semivariogram}). 
$\maxzero{x}=\max(0,x)$ is a rectifier (hinge) function.
$\mrfdistance^2(i,j)$ is a goodness-of-fit statistic~\citep{panaretos2019statistical}. 
}

\rebut{This proposed loss has the following properties: 
}
\rebut{
\vspace{-0.2cm}
\begin{itemize}
\item $\mrfdistance^2(i,j)$ is a goodness-of-fit test statistic. Thus, $\mrfdistance^2(i,j)$ being smaller than certain threshold implies observations $i$ and $j$ can pass the goodness-of-fit hypothesis test under certain significance level. 
\item $\empE_{i^{'},j^{'}\in \mathbb{N}} \mrfdistance^2(i,j) - \delta^0$ and $\empE_{i^{'},j^{'}\in \mathbb{N}_{i,j}} \mrfdistance^2(i,j) - \delta$ can be seen as two thresholds based on the average test statistics plus a desired significance level. 
In the case of Wasserstein-2 distance \citep{panaretos2019statistical}, the test statistics follows a normal distribution with the square of Wasserstein-2 distance of the true underlying models as the mean. We do not have access to the true means, so we use empirical means as estimations.
While $\empE_{i^{'},j^{'}\in \mathbb{N}} \mrfdistance^2(i,j) - \delta^0$   is independent to metric autocorrelation,  $\empE_{i^{'},j^{'}\in \mathbb{N}_{i,j}} \mrfdistance^2(i,j) - \delta$ is dependent. We call them \textit{non-metric threshold} and \textit{metric threshold}, respectively.
By definition, $\empE_{i^{'},j^{'}\in \mathbb{N}_{i,j}} \mrfdistance^2(i,j)$ is exactly double of the empirical generalized model-based semivariogram defined in Section~\ref{sec:general-semivariogram}.
%
\item The rectifier (hinge) function avoids assigning negative values to observation pairs that pass the test, which increases computational stability. 
The idea is similar to the idea of margin and hinge loss in SVMs. 
Our ablation study (Appendix~\ref{sec:ablation}) shows that this choice increases computational stability and clustering accuracy. 
\end{itemize}
\vspace{-0.2cm}
}

\rebut{
Since both $\empE_{i^{'},j^{'}\in \mathbb{N}} \mrfdistance^2(i,j)$ and $\delta^{0}$ are constants, for computational efficiency we can simplify Equation \ref{eq:mcmb-as-test} by defining  $\delta^{0}=\empE_{i^{'},j^{'}\in \mathbb{N}} \mrfdistance^2(i,j)$ and $r(i,j) =  \maxzero{ \mrfdistance^2(i,j) - (\empE_{i^{'},j^{'}\in N_{i,j}} \mrfdistance^2(i,j) - \delta)}$, and rewrite our loss as}
{ 
\begin{equation}\label{eq:mcmb-as-test-simp}
\begin{split}
\modelobjfun(\clusassign) 
 = \sum_{\{C_k \in \clusassign\}}
 \sum_{\{i,j\in C_k\}} [\mrfdistance^2(i,j) + \beta r(i,j)]
\end{split}
\end{equation}
}
\rebut{
which is exactly Equation \ref{eq:mcmb}. This demonstrates that by choosing appropriate $\mbdistance$ and $r(i,j)$, we can theoretically formulate 
clustering as minimizing the penalty for the intra-cluster pairs that do not pass goodness-of-fit tests. This is the central formula that our algorithm is based on.} 

\subsection{Model-based Clustering via \modelfull{} (\loss)}\label{sec:lmcc}

The \loss{} algorithm is \rebut{naturally} derived from incorporating the properties of metric autocorrelation with goodness-of-fit tests. \rebut{As we have discussed in Section \ref{sec:mcmb-good-of-fit}, the penalty function can be defined as}
\begin{equation} \label{eq:lmcc-pre-2}
    \metricpen (i,j) = \maxzero{ \mrfdistance^2(i, j) - [\gamma_m(\metricdis(i,j)) - \delta]}
\end{equation}
Notice we replace the empirical generalized model-based semivariance $\hat{\gamma}_m = \empE_{i^{'},j^{'}\in N_{i,j}} \mrfdistance^2(i,j)$ with the fitted theoretical model-based semivariance $\gamma_m$, because the fitted semivariogram is smooth. In addition, we need to further condition the penalty function on the fitted range $\rho$ (Equation \ref{eq:lmcc}). 
The reason is that when  
$d_c>\rho$, the second term in Eq~\ref{eq:mcmb-as-test-simp} is much smaller than the first term and can be ignored to spare computation. Figure \ref{fig:range-condition} shows an empirical analysis on the iNaturalist-2018 dataset. The average contribution of the penalty function to the total loss beyond $\rho$ quickly drops below $15\%$ and remains flat, which is non-informative and can be omitted in practice.
\rebut{We empirically verified in Appendix \ref{sec:ablation-range-condition} that using the conditional form of penalty function is both beneficial for sparing computation and improving clustering performance.} Thus, the final form of the penalty function is defined as
\begin{equation} \label{eq:lmcc}
    \metricpen_{\rho}(i,j) = \left\{\begin{array}{lr}
        \metricpen(i,j), & \metricdis(i,j) \leq \rho\\
        0, & \metricdis(i,j) > \rho\\
        \end{array}\right\}
\end{equation}
Intuitively, $r_{\rho}$ penalizes observations whose semivariance lies above the theoretical semivariogram. It is \textbf{local} (only effective within the range $\rho$), \textbf{monotonically decreasing}, and \textbf{continuous} (because the semivariogram is monotonically increasing and continuous within the range). 

\begin{figure*}[t!]
\centering
\subfloat[semivariogram with $n=15$ ]{\includegraphics[width = 0.33\textwidth]{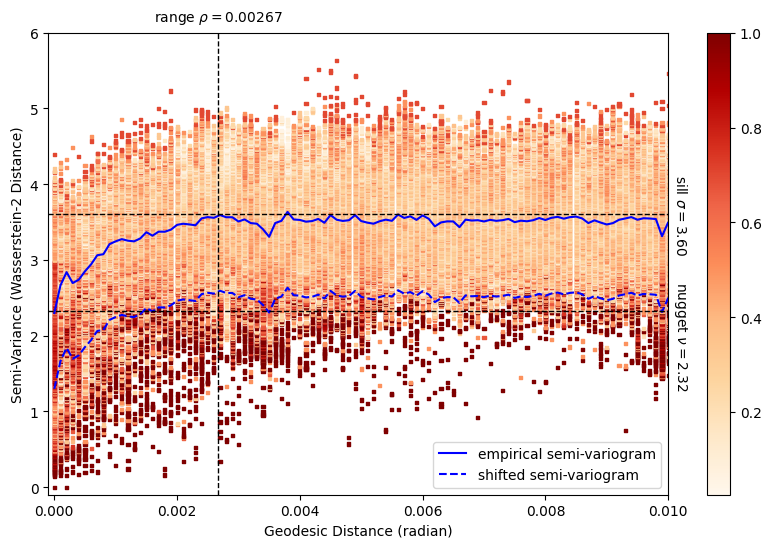}\label{fig:variogram-15}}
\subfloat[semivariogram with $n=20$ ]{\includegraphics[width = 0.33\textwidth]{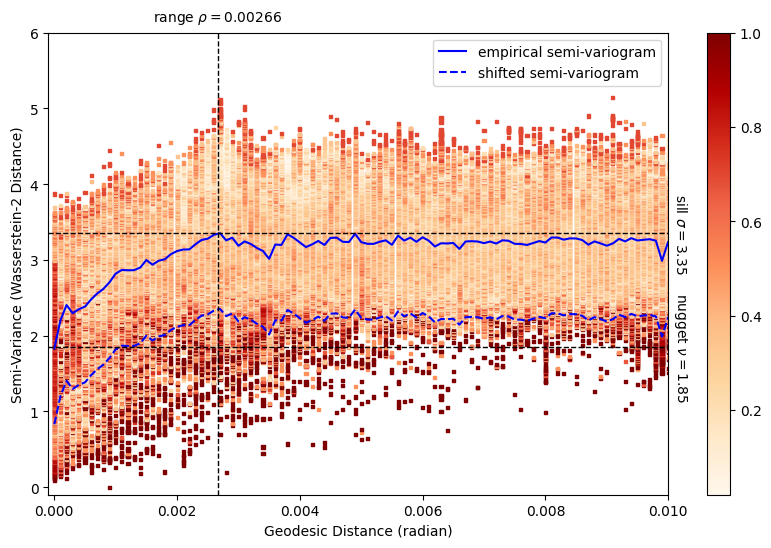}\label{fig:variogram-20}}
\subfloat[semivariogram with $n=30$ ]{\includegraphics[width = 0.33\textwidth]{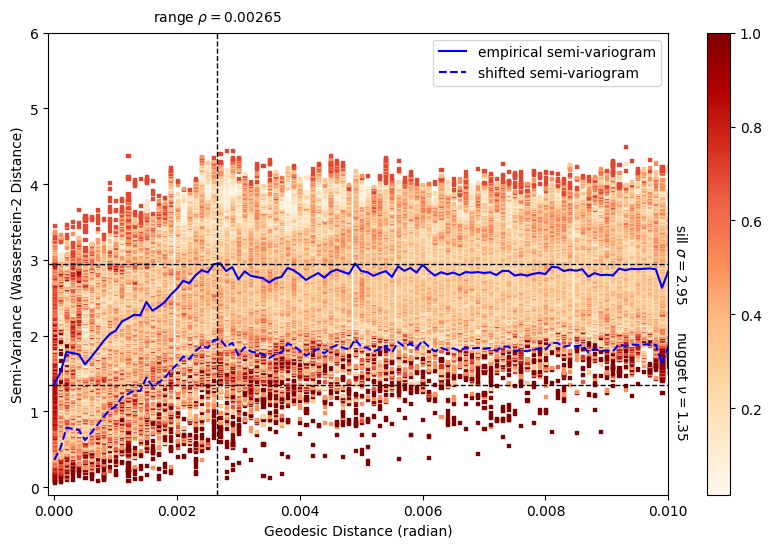}\label{fig:variogram-30}}
\vspace{-0.2cm}
\caption{Empirical generalized model-based semivariogram under different hyperparameter settings. $n$ is the number of nearest neighbors used for fitting the GMRF models for each observation. The color represents the percentage of observation pairs that belong to the same ground-truth cluster in each $0.0001\text{(geodesic)} \times 0.01\text{(Wasserstein-2)}$ bin.}
\label{fig:variogram}
\vspace{-0.3cm}
\end{figure*}

$\delta$ is a critical hyperparameter called \textit{margin}. The motivation of using this hyperparameter comes from both theoretical analysis and empirical observations. \rebut{Theoretically, the semivariogram is an average measurement of intra-cluster model dissimilarity.
Inter-cluster pairs may also happen to have smaller than average semivariance.
To reduce the chance of wrongly identifying inter-cluster pairs as intra-cluster, only observations whose semivariance is significantly small (at least $\delta$ below the theoretical semivariogram) are exempt from penalty. This is equivalent to shifting the 
semivariogram downwards by $\delta$.}
Empirically, by plotting the percentage of ground-truth intra-cluster pairs in each bin, Figure \ref{fig:variogram} shows there is a clear boundary between the dark region (with higher percentage of intra-cluster pairs) and the light region (with lower percentage of intra-cluster pairs). 
We notice that by 
vertically shifting down for an appropriate distance $\delta$, the semivariogram partially overlaps with the boundary. Then penalizing the observation pairs above the shifted semivariogram becomes practically equivalent to penalizing the observation pairs for falling into the region of low intra-cluster probability. Since we do not know the ground-truth value of $\delta$, it is a hyperparameter that needs to be tuned. Algorithm~\ref{alg:LMCC-MC} shows the core \model{} algorithm which is implemented based on Equation \ref{eq:generalized-emp-variogram},  \ref{eq:mcmb-as-test-simp} and \ref{eq:lmcc}.

\begin{algorithm}[h!]
\SetKwInOut{Input}{Input}
\SetKwInOut{Output}{Output}
\Input{
    A dataset $\dataset$ of $\obsnum$ observations $\{\obs_i = (\feavec_i\in F, \posvec_i \in M)\}_{i=1}^\obsnum$. 
    The distance function $d_m$. 
    The metric penalty function $r$. 
    The model fitting algorithm $\mathbf{GL}$. 
    The density-based clustering algorithm $\mathbf{DB}$. 
    The number of neighbors $n$ used for model \rebut{fitting}. 
    The metric-constraint strength $\beta$. 
    The margin hyperparameter $\delta$.
}
\Output{A clustering $\clusassign=\{C_k\}_{k=1}^K$}

for each observation $X_i \in \dataset$ \; 

\:\:\:  find $n$ nearest  observations $\mathbb{N}_i$ in the metric space\;

\:\:\: fit the model parameters $\theta_i \gets \mathbf{GL}(\mathbb{N}_i)$ \hfill (Sec \ref{sec:general-semivariogram})\;

for each pair of observations, compute their \;

\:\:\: model dissimilarity $\mbdistance(i,j) \gets \mrfdistance^2(\theta_i,\theta_j)$  \hfill (Eq \ref{eq:w2})\;

\:\:\: metric distance $\metricdis(i,j) \gets \metricdis(\posvec_i,\posvec_j)$ \;

compute empirical generalized semivariogram $\hat{\gamma}_m$ \hfill  (Eq \ref{eq:generalized-emp-variogram})\;

fit theoretical generalized semivariogram $\gamma_m$ from $\hat{\gamma}_m$\;

compute range $\rho$\ from $\gamma_m$  \hfill (Sec \ref{sec:gsemivariogram})\;

compute 
loss matrix $M^w_{i,j} \gets \mbdistance(i,j) + \beta r_{\rho}(i,j)$ (Eq \ref{eq:lmcc})\;

run density-based clustering algorithm $\clusassign \gets \mathbf{DB}(M^w)$\;

\Return $\clusassign$
\caption{\model{} Algorithm
}
\label{alg:LMCC-MC}
\end{algorithm}
\vspace{-0.3cm}

\section{Experiments}

We perform extensive experiments on two synthetic and seven real-world datasets which cover both temporal and spatial clustering tasks. 
We compare \model~
with a wide range of baselines.
The detailed experiment setup, baseline algorithms and evaluation metrics can be found in Appendix \ref{sec:experiment-setup}.
We conduct hyperparameter tuning on the number of neighbors $n$, the weight $\beta$, and the margin $\delta$. Results show that the search spaces of all hyperparameters have good convexity. Moreover, unlike TICC and STICC, we do not need to re-compute the covariance metrics during hyperparameter tuning for \model{}. That makes hyperparameter tuning for \model{} much faster (see more details in Appendix \ref{sec:tune}). 

\begin{table*}[t!]
\vspace{-0.2cm}
  \caption{Performance on 1-D (temporal) and 2-D (spatial) tasks.
  $d$ denotes the feature dimension, $c$ denotes the ground-truth cluster number, and $N$ denotes the size of each dataset.
  MC stands for metric constraint.
  \textbf{Bold} numbers and \underline{underlined} numbers indicate the best and second best performances. (S)TICC means applying TICC to temporal datasets and STICC to spatial datasets.
  $\beta=0$ means there is no temporal/spatial penalty term applied. - means the method is not suitable for this dataset. NC means the algorithm does not converge.
  \loss-wo/\loss-w represents \loss{} loss without/with metric constraints respectively. 
  }
  \label{tab:baseline}
  \centering 
  \scriptsize 
  \setlength{\tabcolsep}{2.8pt}
  {
  \begin{tabular}{cc|*{4}{c}|*{6}{c}|*{8}{c}}
    \toprule
    \midrule
    \multirow{6}{*}{
    } & \multirow{6}{*}{
    } & \multicolumn{4}{c|}{Synthetic Datasets} & \multicolumn{12}{c}{Real-world Datasets} \\ 

    & & & & & & \multicolumn{6}{c}{Temporal} &  \multicolumn{8}{c}{Spatial} \\

    \cmidrule{3-20}
    & & \multicolumn{2}{c}{\begin{tabular}{@{}c@{}}Temporal \\ \begin{tabular}{@{}c@{}}$d$=5, $c$=5 \\ N=1,000 \end{tabular} \end{tabular}} & \multicolumn{2}{c|}{\begin{tabular}{@{}c@{}}Spatial \\ \begin{tabular}{@{}c@{}}$d$=5, $c$=5 \\ N=10,000 \end{tabular} \end{tabular}} & \multicolumn{2}{c}{\begin{tabular}{@{}c@{}}Pavement \\ \begin{tabular}{@{}c@{}}$d$=10, $c$=3 \\ N=1,055 \end{tabular} \end{tabular}} & \multicolumn{2}{c}{\begin{tabular}{@{}c@{}}Vehicle \\ \begin{tabular}{@{}c@{}}$d$=7, $c$=5 \\ N=16,641 \end{tabular} \end{tabular}} &\multicolumn{2}{c|}{\begin{tabular}{@{}c@{}}Gesture \\ \begin{tabular}{@{}c@{}}$d$=3, $c$=8 \\ N=704,970 \end{tabular} \end{tabular}} & \multicolumn{2}{c}{\begin{tabular}{@{}c@{}}Climate \\ \begin{tabular}{@{}c@{}}$d$=5, $c$=14 \\ N=4,741 \end{tabular} \end{tabular}} & \multicolumn{2}{c}{\begin{tabular}{@{}c@{}}iNat2018 \\ \begin{tabular}{@{}c@{}}$d$=16, $c$=6 \\ N=24,343 \end{tabular} \end{tabular}} & \multicolumn{2}{c}{\begin{tabular}{@{}c@{}}POI \\ \begin{tabular}{@{}c@{}}$d$=7, $c$=10 \\ N=23,019 \end{tabular} \end{tabular}} & \multicolumn{2}{c}{\begin{tabular}{@{}c@{}}Landuse \\ \begin{tabular}{@{}c@{}}$d$=7, $c$=5 \\ N=8,964 \end{tabular} \end{tabular}}\\
    \cmidrule{3-20}
    Model Type & Model & ARI & NMI & ARI & NMI & ARI & NMI & ARI & NMI & ARI & NMI & ARI & NMI & ARI & NMI & ARI & NMI & ARI & NMI\\
    \midrule
    \multirow{4}{*}{\begin{tabular}{@{}c@{}} No-Constraint \\ Model-Free  \end{tabular}} & k-Means & 1.03 & 1.69 & 1.26 & 1.66 & 8.02 & 6.59 & 8.94 & 21.54 & 2.78 & 5.23 & 5.47 & 22.14 & 6.91 & 14.71 & 18.37 & 43.44 & 2.39 & 4.21 \\
    & DBSCAN & 2.44 & 2.50 & 3.69 & 5.38 & 15.25 & 18.75 & 33.67 & 41.83 & 1.18 & 2.07 & 3.61 & 17.89 & 34.91 & 34.69 & 15.03 & 39.29 & 11.91 & 7.19 \\
    & HDBSCAN & 0.90 & 0.61 & 1.00 & 1.39 & 7.10 & 11.66 & 37.51 & 41.64 & - & - & 11.52 & 28.01 & 7.65 & 17.92 & 20.78 & 62.55 & 1.00 & 7.64 \\
    & DTW & 2.52 & 2.13 & - & - & 17.13 & 17.55 & 8.11 & 23.35 & - & - & - & - & - & - & - & - & - & - \\
    \midrule
    \multirow{3}{*}{\begin{tabular}{@{}c@{}} Constrained \\ Model-Free  \end{tabular}} & PCK-Means & 5.12 & 5.68 & 2.30 & 2.89 & 7.42 & 5.13 & 4.80 & 14.17 & NC & NC & 18.50 & 34.67 & \underline{25.51} & 28.96 & 0.12 & 0.18 & 0.11 & 0.26 \\
    & MDST-DBSCAN & - & - & 1.12 & 5.73 & - & - & - & - & - & - & 11.32 & 27.89 & 8.43 & 18.13 & 1.33 & 0.97 & 1.29 & 1.01 \\
    & SKATER & - & - & 23.87 & 32.29 & - & - & - & - & - & - & \textbf{23.44} & \textbf{44.10} & 0.51 & 0.35 & 1.52 & 0.91 & 1.03 & 0.74  \\
    \midrule
    \multirow{3}{*}{\begin{tabular}{@{}c@{}} No-Constraint \\ Model-Based  \end{tabular}} & GMM & 7.82 & 9.54 & 9.26 & 10.35 & 28.05 & 28.74 & 57.87 & 58.78 & 2.44 & 4.15 & 19.06 & 34.97 & 21.72 & 35.91 & 16.38 & 42.96 & 2.86 & 4.61 \\
    & (S)TICC-$\beta$=0 & 80.11 & 83.95 & 91.28 & 89.28 & 58.54 & 58.83 & 40.12 & 45.86 & 3.26 & 6.56 & 13.30 & 30.53 & NC & NC & 13.29 & 27.08 & 7.22 & 12.60 \\
    & \model-wo & \underline{86.38} & 84.56 & 87.34 & 84.74 & \underline{76.10} & \underline{74.36}& \underline{63.31} & \underline{58.60} & 8.12 & \underline{33.60} & 16.63 & 36.73 & 21.90 & \underline{36.47} & \underline{30.45} & \underline{66.23} & \underline{12.91} & \underline{28.72} \\
    \midrule
    \multirow{2}{*}{\begin{tabular}{@{}c@{}} Constrained \\ Model-Based  \end{tabular}} & (S)TICC & 84.88 & \underline{86.13} & \underline{91.84} & \underline{89.85} & 62.27 & 61.89 & 50.53 & 53.68 & \underline{12.20} & 23.20 & 17.62 & 37.29 & NC & NC & NC & NC & 11.04 & 15.35 \\
    & \model-w & \textbf{90.50} & \textbf{87.96} & \textbf{94.49} & \textbf{91.98} &\textbf{77.64} & \textbf{77.22} & \textbf{65.04} & \textbf{59.36} & \textbf{26.51} & \textbf{55.34} & \underline{20.08} & \underline{40.91} & \textbf{42.70} & \textbf{40.49} & \textbf{39.81} & \textbf{68.27} & \textbf{36.54} & \textbf{42.97} \\
    \hline
    \hline
  \end{tabular}
  }
  \vspace{-0.2cm}
\end{table*}

\subsection{Main Result}

From Table \ref{tab:baseline} we can see in general, model-based algorithms handle spatio-temporally distributed data better than feature-based clustering algorithms. Our method (\loss-w) outperforms the strong baselines (TICC and STICC) in all tasks. 
Besides performance improvement, \model~ is also more flexible and generally applicable. \model~ performs consistently well throughout different constraint dimensions, dataset sizes and cluster numbers, whereas TICC and STICC can only handle either 1D or 2D metric constraints and do not converge stably (NC in Table \ref{tab:baseline}), especially when the feature dimension and the dataset size are large. Furthermore, comparing TICC, STICC and \loss-w with their non-constrained version (i.e., TICC ($\beta=0$), STICC ($\beta=0$) and 
\loss-wo), 
we can see that metric constraints do improve the clustering quality.

One important observation is that SKATER~\citep{assunccao2006efficient} performs even better than \model{} on the Climate dataset, but works extremely bad on the iNaturalist-2018/POI/Land-use datasets. There are two take-aways: (1) The ground-truth clusters of Climate dataset are contiguous regions with disjoint boundaries, whereas those of the latter three datasets overlap with each other (Figure \ref{fig:ground-truth-shape}). SKATER splits the metric space into a Voronoi diagram, so it fits the former dataset well but fails the latter. It is not as generally applicable as \model{}. (2) The performance of \model{} is worse on the Climate dataset because the observations are spatially sparse, e.g., the maximum distance between 30-nearest neighbors may be as large as 1 radian. 
This brings a dilemma: in order to have enough samples to estimate the underlying models, we must risk including metrically distant observations, which by the metric autocorrelation, have high variance. This is a limitation of our algorithm. 
See Appendix~\ref{sec:more-exp-discussions} for more discussions of the result.

\subsection{Stability, Robustness and Efficiency}
\model{} is robust and computationally stable in two ways. Firstly, it is a sequential algorithm. TICC/STICC, on the other hand, uses EM iterations which may accumulate error. For example, we observe that if the initial cluster assignment is too imbalanced, TICC/STICC will self-enhancingly increase this imbalance until it fails to converge. Secondly, \model{} has only three hyperparameters to tune: the number of neighbors $n$; the penalty weight $\beta$ and the shift $\delta$. $\beta$ and $n$ are common hyperparameters that all model-based clustering algorithms share. Thus, only the shift $\delta$ is unique to our method. Furthermore, by comparing Figure \ref{fig:variogram-15}, \ref{fig:variogram-20} and \ref{fig:variogram-30}, we find that the key factors of the semivariogram (range, sill and nugget) remain relatively stable. This finding is critical because our method is heavily based on the reliable construction of the semivariogram.

\model~ is more efficient than TICC/STICC by removing EM iterations.
The underlying model of each observation is only estimated once throughout the entire algorithm of \model~, whereas TICC/STICC must re-estimate models in every iteration. Empirically, the execution speed of our method is 5 to 15 times faster than TICC/STICC. Moreover, the estimated underlying models can be archived and reused, making hyperparameter tuning much easier than TICC/STICC. 
Please see Appendix \ref{sec:runtime} for a detailed comparison of theoretical and empirical runtime complexity between \model{} and TICC/STICC. 



\subsection{Ablation Studies}
\label{sec:ablation}


\rebut{We conduct a series of ablation studies on the Pavement dataset to investigate the necessity and effectiveness of the components we adopt in our algorithm.}

\subsubsection{Wasserstein-2 Distance vs other feature similarity measures}
\label{sec:abla-sim}

To demonstrate how the choice of different feature similarity measures matters in clustering, we replace the Wasserstein-2 distance in Equation \ref{eq:generalized-emp-variogram} with various model-free/model-based measures and report the experiment results on the Pavement dataset in Table \ref{tab:ablation-wasser}. We can see that the model-based Wasserstein-2 distance significantly outperforms both the model-free and the model-based alternatives.

\begin{table}[ht!]
\vspace{-0.2cm}
  \centering 
  \caption{\rebut{Comparing different feature similarity measures}}  \label{tab:ablation-wasser}
  \setlength{\tabcolsep}{2pt}
  {\scriptsize 
  \begin{tabular}{*{13}{c}}
    \toprule
    \midrule
    & \multicolumn{2}{c}{Wasserstein-2}
    & \multicolumn{2}{c}{Euclidean}
    & \multicolumn{2}{c}{Cosine}
    & \multicolumn{2}{c}{Total Var.}
    & \multicolumn{2}{c}{KL-D}
    & \multicolumn{2}{c}{JS-D} \\
    \midrule
    & ARI & NMI & ARI & NMI & ARI & NMI & ARI & NMI & ARI & NMI & ARI & NMI\\
    & 77.64 & 77.22 & 23.11 & 22.43 & 0.34 & 1.36 & 3.37 & 3.61 & 56.73 & 66.10 & 15.55 & 18.84\\
    \hline
    \hline
  \end{tabular}
  }
\vspace{-0.8cm}
\end{table}

\begin{figure*}[ht!]
\centering
\vspace{-0.2cm}
\subfloat[\rebut{Cosine distance}]{\includegraphics[width = 0.3\textwidth]{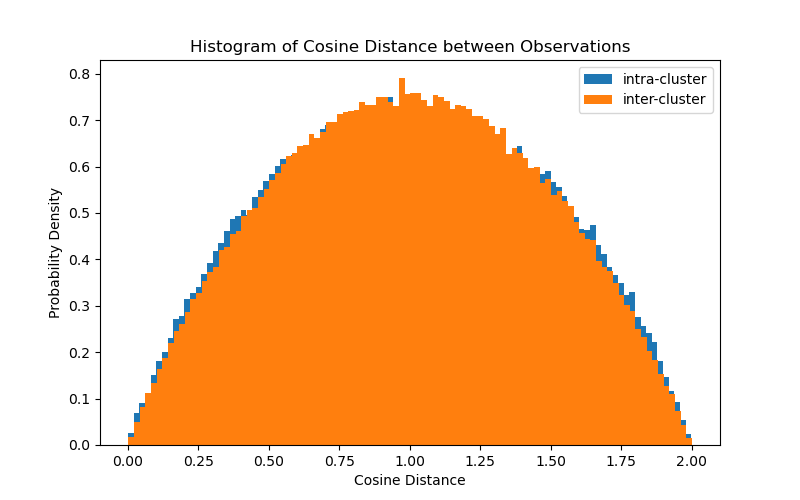}\label{fig:cosine-hist}}
\subfloat[\rebut{Euclidean distance}]{\includegraphics[width = 0.3\textwidth]{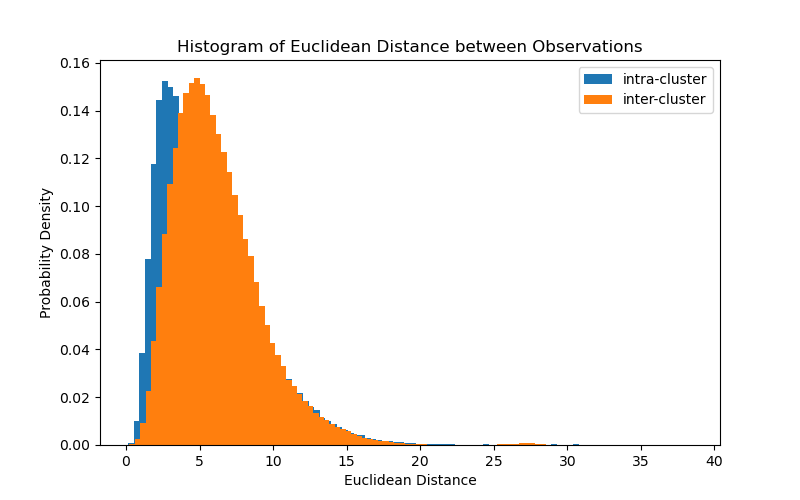}\label{fig:euclidean-hist}}
\subfloat[\rebut{Wasserstein-2 distance}]{\includegraphics[width = 0.3\textwidth]{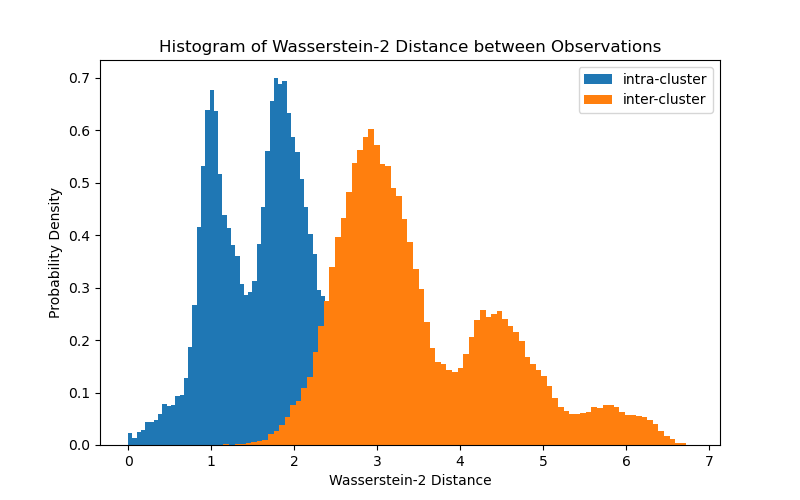}\label{fig:wasserstein-hist}}
\vspace{-0.2cm}
\caption{\rebut{The histograms of pairwise distance between intra-cluster and inter-cluster observations in the Pavement dataset.
The distributions of intra/inter-cluster Wasserstein-2 distance show more distinctive patterns than those of cosine distance and Euclidean distance.
}}
\label{fig:ablation-feature-distance}
\vspace{-0.4cm}
\end{figure*}

Here we present a simple experiment to show how a model-based approach outperforms model-free distance when the underlying model is properly selected. Figure~\ref{fig:ablation-feature-distance} depicts the histograms of pairwise Wasserstein-2 distance, Euclidean distance, and cosine distance between two observations in the Pavement dataset. 
The blue represents intra-cluster pairs and the orange represents inter-cluster pairs. We can see the model-based Wasserstein-2 distance itself makes distinguishing intra-cluster and inter-cluster pairs a lot easier than using Euclidean or cosine distance. It clearly demonstrates that raw feature similarity measures can not capture the complex patterns. Besides, Euclidean distance is known to be inefficient in high dimensions due to sparsity~\citep{aggarwal2001surprising}, and the cosine distance is unable to represent differences in magnitudes.

Then we analyze why the other model-based measures yield inferior results. Total variation and Jensen-Shannon (JS) divergence perform poorly because they are difficult to accurately compute in high-dimensional spaces. KL-divergence underperforms our Wasserstein-2 distance because it is less sensitive to fine-grained model differences.

\subsubsection{GLasso vs other covariance estimaters}
\label{sec:abla-cov}

\rebut{We use Graphical Lasso 
as the covariance estimation algorithm, but the effectiveness of \model{} does not rely on this specific implementation. As an ablation study, we replace Graphical Lasso with Minimum Covariance Determinant (MIN-COV)\footnote{https://scikit-learn.org/stable/modules/generated/sklearn.covariance.MinCovDet.html} and Shrunk Covariance (SHRUNK)\footnote{https://scikit-learn.org/stable/modules/generated/sklearn.covariance.ShrunkCovariance.html} and apply \model{} on the Pavement dataset. Clustering performance is reported in Table \ref{tab:ablation-glasso}. All results are under the best hyperparameters based on grid search.}
\rebut{In general, the more robust and more accurate the covariance estimation algorithm is, the better the clustering performance is. Shrunk is the least robust covariance estimation algorithm among the three, thus its performance is obviously lower than GLasso and MinCov. However, different variations of \model{} still significantly outperform the strongest baseline, TICC. It indicates the effectiveness of \model{}.}

\vspace{-0.4cm}
\begin{table}[ht!]
  \centering 
  \caption{\rebut{Comparing different covariance estimation methods}}  \label{tab:ablation-glasso}
  \setlength{\tabcolsep}{4pt}
  {\scriptsize
  \begin{tabular}{*{9}{c}}
    \toprule
    \midrule
    \multirow{2}{*}{Method} 
    & \multicolumn{2}{c}{TICC}
    & \multicolumn{2}{c}{\model}
    & \multicolumn{2}{c}{\model}
    & \multicolumn{2}{c}{\model}\\
    & \multicolumn{2}{c}{{(Baseline)}}
    & \multicolumn{2}{c}{{(GLASSO)}}
    & \multicolumn{2}{c}{{(MIN-COV)}}
    & \multicolumn{2}{c}{{(SHRUNK)}}\\
    \midrule
    \multirow{2}{*}{Performance} 
    & ARI & NMI & ARI & NMI & ARI & NMI & ARI & NMI \\
    & 62.27 & 61.89 & 77.64 & 77.22 & 80.82 & 73.78 & 74.70 & 71.42\\
    \hline
    \hline
  \end{tabular}
  }
\vspace{-0.4cm}
\end{table}

\subsubsection{DBSCAN vs other distance-based clustering algorithms} 
\label{sec:abla-cluster}
\rebut{Similarly, \model{} does not rely on any specific implementation of the clustering algorithm. Table \ref{tab:ablation-dbscan} again demonstrates that though clustering performances are affected by the choice of distance-based clustering algorithms, \model{} still outperforms the baselines by large margins.}

\vspace{-0.4cm}
\begin{table}[ht!]
  \centering 
  \caption{\rebut{Comparing different distance-based clustering algorithms}}  \label{tab:ablation-dbscan}
  \setlength{\tabcolsep}{4pt}
  {\scriptsize
  \begin{tabular}{*{9}{c}}
    \toprule
    \midrule
    Method
    & \multicolumn{2}{c}{{TICC}}
    & \multicolumn{2}{c}{{\model} }
    & \multicolumn{2}{c}{{\model} }
    & \multicolumn{2}{c}{{\model} }\\
    & \multicolumn{2}{c}{{ (Baseline)} }
    & \multicolumn{2}{c}{{(DBSCAN)} }
    & \multicolumn{2}{c}{{(HDBSCAN)} }
    & \multicolumn{2}{c}{{(OPTICS)} }\\
    \midrule
    \multirow{2}{*}{Performance} & ARI & NMI & ARI & NMI & ARI & NMI & ARI & NMI \\
    & 62.27 & 61.89 & 77.64 & 77.22 & 72.35 & 69.61 & 69.77 & 68.58\\
    \hline
    \hline
  \end{tabular}
  }
\vspace{-0.4cm}
\end{table}

\subsubsection{The hinge operation in the loss function} 
\label{sec:abla-hinge}

\rebut{In our ablation experiment, removing the rectifier (hinge) operation causes computational instability. 
Since we apply a distance-based clustering algorithm on top of the weighted distance matrix, all entries are required to be positive. When we remove the max operation, the weighted distance sometimes becomes negative and the clustering algorithm fails.}

\subsubsection{The range condition in the penalty}
\label{sec:ablation-range-condition}

\rebut{As an ablation study, we ignore the range condition in Equation \ref{eq:lmcc}, i.e., define $r_{\rho}(i,j)$ simply as}
\begin{equation} \label{eq:lmcc-no-range}
    \metricpen_{\rho}(i,j) = \maxzero{ \mrfdistance^2(i, j) - [\gamma_m(\metricdis(i,j)) - \delta]}
\end{equation}
{This means we need to compute the penalty term for all possible pairs of data points. We then apply \model{}. The comparison of clustering performance using Equation \ref{eq:lmcc} (Conditional) and using Equation \ref{eq:lmcc-no-range} (Unconditional) is demonstrated in Table \ref{tab:ablation-penalty-condition}. Ignoring the range does not only negatively affects the clustering performance, but also wastes resources, since we can spare the computation of penalty terms of pairs out of range.}

\vspace{-0.4cm}
\begin{table}[ht!]
  \centering 
  \caption{\rebut{Comparing conditional and unconditional penalty }}  \label{tab:ablation-penalty-condition}
  \setlength{\tabcolsep}{2pt}
  {\scriptsize
  \begin{tabular}{*{5}{c}}
    \toprule
    \midrule
    Method
    & \multicolumn{2}{c}{\pboxtwo{\loss}{(Conditional)}} 
    & \multicolumn{2}{c}{\pboxtwo{\model}{(Unconditional)}} \\
    \midrule
    \multirow{2}{*}{Performance} & ARI & NMI & ARI & NMI \\
    & 77.64 & 77.22 & 76.91 & 76.25 \\
    \hline
    \hline
  \end{tabular}
  }
\vspace{-0.4cm}
\end{table}

\section{Conclusion and Future Works}

In this paper, we propose a novel clustering technique called \model~that injects knowledge of metric autocorrelation into model-based clustering algorithms by computing pairwise Wasserstein-2 distance between estimated model parameterizations for each observation. 
\model~ provides a unified solution to clustering problems with temporal/spatial/higher-dimensional metric constraints and achieves SOTA performance on both synthetic and real-world datasets. 
Moreover, by minimizing the total hinge loss of pairwise goodness-of-fit tests,
\model~ is more computationally efficient and stable than the strongest baselines TICC and STICC, which optimize data likelihood through EM procedures during clustering. 

For future work, it is worth extending \model~ to non-Gaussian, general Markov Random Fields \rebut{using their corresponding pairwise Wasserstein-2 distances}.

\newpage

\section*{Impact Statement}

This paper presents work whose goal is to advance the field of Machine Learning. We do not foresee any potential negative societal impacts of the current work.

\section*{Software and Data}
All the datasets and packages we use are publicly accessible online and properly cited.
The implementation of our algorithm and a tutorial is publicized on GitHub via \url{https://github.com/Octopolugal/MC-GTA.git}.

\section*{Acknowledgements} 

This work is mainly funded by the National Science Foundation under Grant No. 2033521 A1 – KnowWhereGraph: Enriching and Linking Cross-Domain Knowledge Graphs using Spatially-Explicit
AI Technologies. Gengchen Mai acknowledges support from the Microsoft Research Accelerate Foundation Models Academic Research (AFMR) Initiative. Any opinions, findings, and conclusions, or recommendations expressed in this material are those of the authors and do not necessarily reflect the views of the National Science Foundation.



\newpage
\appendix
\onecolumn

\section{Appendix} \label{sec:app}

\subsection{A General Objective of Clustering}  \label{sec:app_general_cluster_obj}
While the complete definition of clustering has not yet come to an agreement, three principles in general apply~\citep{jain1988clustering}: (1) \textbf{Intra-Cluster Cohesion}: observations, in the same cluster, must be as similar as possible; (2) \textbf{Inter-Cluster Separation}: observations, in different clusters, must be as different as possible; (3) \textbf{Interpretability}: measurement for similarity and dissimilarity must be clear and have practical meanings.
Following these principles, the two key components of a clustering algorithm are the similarity/dissimilarity measurement and the algorithm that optimizes intra-cluster cohesion and inter-cluster separation. There is no cure-all clustering algorithm. Different data structures require different similarity/dissimilarity measurements (e.g. cosine distance, Euclidean distance, graph distance) and different optimization algorithms (e.g. hierarchical, iterative, estimation-maximization-based), resulting in a variety of clustering algorithms such as partition-based clustering, hierarchical clustering, density-based clustering, model-based clustering, etc. 

Conceptually, if we define measurements similarity $s(\cdot, \cdot)$ and dissimilarity $d(\cdot, \cdot)$ between two observations $\obs_i, \obs_j$, we can view a clustering problem as finding an optimal cluster assignment $\hat{C}_K$ that maximizes the objective:
\begin{align}
\begin{split}
    \hat{C}_K & 
    =\text{ arg max}_{C} \Big[  
     \sum_{\{C_k \in C\}}
     \sum_{\{i,j\in C_k\}} s(X_i,X_j)
    + \beta \sum_{\{C_k, C_l \in C, k \neq l\}}\sum_{\{i\in C_k, j \in C_l\}}d(X_i,X_j)\Big]
\end{split}
\end{align}

The first term corresponds to the intra-cluster cohesion principle and the second term corresponds to the inter-cluster separation principle. $\beta$ is a hyperparameter chosen to control how much we weigh these two terms, 
since the two objectives may compete. The choice of $s$ and $d$, in turn, corresponds to the interpretability principle. Usually we simply let $s(\cdot, \cdot) = -d(\cdot, \cdot)$, thus the objective becomes
\begin{align}
\begin{split}
    \hat{C}_K &=\text{ arg min}_{C} \Big[  
     \sum_{\{C_k \in C\}}
     \sum_{\{i,j\in C_k\}} d(X_i,X_j) - \beta \sum_{\{C_k, C_l \in C, k \neq l\}}\sum_{\{i\in C_k, j \in C_l\}}d(X_i,X_j)\Big]
\end{split}
\end{align}

\subsection{Gaussian Markov Random Field (GMRF)}  \label{sec:gmrf}
A Gaussian Markov Random Field (GMRF) is a special case of the general Markov Random Field (MRF)~\citep{wang2013markov}, which additionally requires the joint and marginal distributions of variables to be Gaussian. Using GMRFs introduces several advantages. The first advantage is high computational efficiency. A (centered) GMRF can be efficiently represented and fitted as a sparse covariance matrix, through Graphical LASSO\footnote{
https://scikit-learn.org/stable/modules/generated/
sklearn.covariance.graphical\_lasso.html}~\citep{10.1093/biostatistics/kxm045}. Secondly, a GMRF can provide interpretable insights into variable correlations. Finally, a GMRF can be used to properly model continuous data in a wide range of situations~\citep{rue2002fitting, hartman2008fast}. For example, in spatial data mining, many commonly used real-valued features, such as place check-in numbers ~\citep{mckenzie2015regional,janowicz2019using,yan2017itdl}, traffic volume ~\citep{liu2017road2vec,cai2020traffic}, customer rating ~\citep{gao2017extracting}, sustainability indices \citep{yeh2021sustainbench,elmustafa2022understanding,manvi2024geollm}, and real-estate pricing ~\citep{law2019take,kang2021understanding}, can be treated as normal distributions after standardization. In addition to that, the covariance representation of a GMRF can be easily extended into a Toeplitz matrix that models inter-observation dependency, which is very important in understanding the interactions across time~\citep{hallac2017ticc} and space~\citep{kang2022sticc}.
Due to the above advantages of GMRF, we choose \rebut{it} as the parametrization of the underlying models in our method. 
\rebut{Furthermore, the other important component of our method, the Wasserstein-2 distance~\citep{gibbs2002choosing}, works best with GMRFs. It is mathematically proved that the Wasserstein-2 distance has a closed-form solution on GMRF models, which ensures the efficiency and stability of our method.}
\subsection{Metricization of Weak Convergence}\label{sec:wasserstein-proof}

To read more detailed discussions of the metrization of probability convergence, see \citet{gibbs2002choosing} for a comprehensive summary. According to the same paper, two important propositions are worth notification: 1) the L\'evy-Prokhorov metric is "precisely the minimum distance 'in probability' between random variables distributed according to $\mu$ and $\nu$", and 2) the L\'evy-Prokhorov metric and the Wasserstein's distance satisfy the following quantitative relation:
\begin{equation}\label{eq:4}
    \pi \leq W_p \leq (\text{diam}(\Omega) + 1) \pi
\end{equation}
where $\text{diam}(\Omega) := \sup \{d(x,y):x, y \in \Omega\}$ is the diameter of the sample space $\Omega$. These two propositions justify that though the Wasserstein's distance is not the tightest bound (i.e., the Prokhorov metric), it converges as fast up to a constant factor, so long as the metric space is bounded.

Since both the L\'evy-Prokhorov metric and the Wasserstein's distance has guaranteed convergence, the choice of $d_m$ is mainly upon computational efficiency. Whereas both metrizations have no simple algorithms for computation in the general case, the Wasserstein-2 distance between two multi-variate Gaussian distributions has a neat closed-form formula in terms of mean vectors and covariance matrices. \citet{https://doi.org/10.1002/mana.19901470121} gives the formula of the squared Wasserstein-2 distance as follows:
\begin{equation}\label{eq:5}
    W_2^2(\mrfparam_1, \mrfparam_2) = d_2^2(\mu_1, \mu_2) + \text{Tr}(\Sigma_1 + \Sigma_2 - 2(\Sigma_1^{1/2}\Sigma_2 \Sigma_1^{1/2})^{1/2})
\end{equation}

Here $\mu_1, \mu_2$ are mean vectors and $\Sigma_1, \Sigma_2$ are covariance matrices. $\mrfparam = (\mu, \Sigma)$. Tr is the trace of a matrix. 

By computing the pairwise Wasserstein-2 distance between the estimated models, we obtain a distance matrix. Any density-based clustering algorithms that support pre-computed distance matrix, such as DBSCAN~\citep{10.5555/3001460.3001507}, can be seamlessly applied without any modification. Since these clustering algorithms are designed to minimize intra-cluster distance and maximize inter-cluster distance, it follows immediately that the intra-cluster observations follow as similar as possible distributions whereas the inter-cluster observations follow as dissimilar as possible distributions, by the fact that the Wasserstein's distance is a metrization of weak convergence. 

With this dimension reduction, we can finally transform the original metric-constrained model-based clustering problem in the product space $F \times M$ to a simpler problem in the product space $\mathcal{R} \times M$. Since $\mathcal{R}$ and $M$ are both metric spaces, density-based algorithms that are supported on product metric spaces such as ST-DBSCAN\citep{BIRANT2007208} can be then applied. However, these algorithms treat the two metric spaces independently without considering the correlation introduced by the metric constraint. In Section \ref{sec:lmcc}, 
we discuss how to address this issue.
\subsection{Definitions of Distance Metrics} \label{app:lpm}
The definitions of L\'evy-Prokhorov Metric and Wasserstein's distance are as follow.

\textbf{L\'evy-Prokhorov Metric}: given a separable metric space $(M, d)$ together with its Borel sigma algebra $\mathcal{B}(M)$, define the $\epsilon$-neighborhood of $A \subset M$ as $A^{\epsilon} := \{p \in M: \exists q \in A \text{ s.t. } d(p,q) < \epsilon\}$. Then the L\'evy-Prokhorov metric $\pi$ of two probability measures $\mu, \nu$ is defined as
\begin{align}\label{eq:2}
\begin{split}    
    \pi (\mu, \nu) := &
    \inf \{\epsilon > 0: \mu(A) \leq \nu(A^{\epsilon}) + \epsilon \text{ and } \nu(A) \leq \mu(A^{\epsilon}) + \epsilon, \forall A \in \mathcal{B}(M) \}
\end{split}
\end{align}

\textbf{Wasserstein's Distance}: given a Radon metric space $(M, d)$, for $p \in [1, \infty)$, the Wasserstein-$p$ distance $W_p$ between two probability measures $\mu, \nu$ is defined as
\begin{equation}\label{eq:3}
    W_p := (\inf_{\gamma \in \Gamma(\mu, \nu)} \mathbb{E}_{(x,y)\sim \gamma d(x,y)^p})^{1/p}.
\end{equation}
Here $\Gamma(\mu, \nu)$ is the set of all possible couplings of $\mu$ and $\nu$.

\subsection{Experiment Setup}\label{sec:experiment-setup}

\subsubsection{Baseline Models and Evaluation Metrics}
\textbf{Baseline Models}. 
We compare our method to both density-based and model-based clustering algorithms. See Table \ref{tab:baseline} for details. Among them, TICC\citep{hallac2017ticc} can only deal with 1-dimensional constraint and STICC\citep{kang2022sticc} can only deal with 2-dimensional constraint. Thus, the former will only be evaluated against 1-dimensional datasets and the latter only against 2-dimensional datasets. All other models that do not incorporate metric constraint information are evaluated on both 1-dimensional and 2-dimensional datasets.

We compare our \model{} with a wide range of baseline clustering algorithms. (1) General non-constrained clustering algorithms: kMeans~\citep{ahmed2020k}, DBSCAN~\citep{10.5555/3001460.3001507}, HDBSCAN~\citep{mcinnes2017hdbscan}.
(2) Must-link/Cannot-link-based constrained clustering algorithms: PCKMeans~\citep{basu2004active})
, which use distance matrix to sample must-links/cannot-links. (3) Temporal/Spatial clustering algorithms: DTW~\citep{yadav2018dynamic} (temporal), MDST-DBSCAN~\citep{ijgi10060391} (the multivariate version of ST-DBSCAN~\citep{birant2007st}, spatial-temporal), SKATER~\citep{assunccao2006efficient} (spatial).
(4) Model-based clustering algorithms GMM~\citep{reynolds2009gaussian}, TICC~\citep{hallac2017ticc} and STICC~\citep{kang2022sticc}. 

\textbf{Evaluation Metrics of Clustering Quality}. For the fairness of comparison, we adopt the most commonly used ground-truth label based metrics, Adjusted Rand Index (ARI) 
\citep{Hubert1985ComparingP} and Normalized Mutual Information (NMI)
\citep{article}. We use their implementation in \texttt{sklearn}\citep{scikit-learn}. We do not adopt the Macro-F1 metric that TICC\citep{hallac2017ticc} uses because this metric is only well-defined when cluster number is fixed, while our method is density-based, which does not preset a cluster number.

\subsubsection{Synthetic Dataset}
We generate 1-dimensional and 2-dimensional synthetic datasets following the \loss{} assumption discussed in Section \ref{sec:lmcc}. The only hyperparameters we preset are cluster number $K$, feature dimension $D$, noise scale $\alpha$ and sample batch size $k$. All other hyperparameters such as sequence length, cluster size and so forth are completely randomly generated for the sake of fair comparison.

\textbf{1-Dimensional Synthetic Dataset}. We generate the 1-dimensional synthetic dataset following the \loss{} assumption discussed in Section \ref{sec:lmcc}:

\setlist[itemize]{leftmargin=5.5mm}
\begin{itemize}[noitemsep,topsep=0pt,parsep=0pt,partopsep=0pt]
    \item Choose hyperparameters: cluster number $K$, feature dimension $D$, noise scale $\alpha$, sample size $k$.
    \item Randomly choose a subsequence number $N$.
    \item Randomly generate $K$ different $D \times D$ ground-truth covariance matrices $\{\Sigma_1, \Sigma_2 \cdots \Sigma_K \}$. It is required that the pairwise Wasserstein-2 distances should all be greater than 1.0. This is to make sure that observations of different clusters are statistically different.
    \item Generate a random list of $N$ ground-truth subsequence cluster labels $\{C_1, C_2 \cdots C_N\}, C_i \in \{0..K-1\}$, and a random list of $N$ subsequence lengths $\{L_1, L_2 \cdots L_N\}$.
    \item For each subsequence label $C_i$ and subsequence length $L_i$, generate $L_i$ perturbed covariance matrices $\{P_{i,1}, P_{i,2} \cdots P_{i,L_i}\}$ by adding Gaussian noise to $\Sigma_i$. Notice, in order to conform with the monotonic assumption, we add noise with noise scale $j\alpha$ as we generate $P_{i,j}$, and the maximum noise scale should be no larger than $10\%$ of the maximum entry in the ground-truth covariance matrix, in order to meet the continuous assumption.
    \item Sample $k$ $D$-dimensional feature vectors from each $P_{i,j}$ sequentially and concatenate them all together into a $k \Sigma_{i=1}^{N}L_i$ list, with each entry being a $D$-dimensional feature vector. The corresponding position list is simply $\{1, 2 \cdots k \Sigma_{i=1}^{N}L_i\}$. Pairing the feature list and the position list makes the dataset.
\end{itemize}

\textbf{2-Dimensional Synthetic Dataset}. We generate the 2-dimensional synthetic dataset, also following the \loss{} assumption discussed in Section \ref{sec:lmcc}:

\setlist[itemize]{leftmargin=5.5mm}
\begin{itemize}[noitemsep,topsep=0pt,parsep=0pt,partopsep=0pt]
    \item Choose hyperparameters: cluster number $K$, feature dimension $D$, noise scale $\alpha$.
    \item Randomly choose a list of cluster sizes $\{N_1 .. N_K\}$.
    \item Randomly generate $K$ points $\{\posvec_1 .. \posvec_K\}$ on the $X-Y$ plane as the metric center of clusters. Randomly generate $K$ $ 2\times 2$ covariance matrices $\{S_1 .. S_K\}$. For each $\posvec_i$, generate $N_i$ points $\{\posvec_{i,1} .. \posvec_{i,N_i}\}$ from the bivariate Gaussian distribution specified by $S_i$. For each generated point, its ground-truth cluster label is $i$. 
    \item Randomly generate $K$ different $D \times D$ ground-truth covariance matrices $\{\Sigma_1, \Sigma_2 \cdots \Sigma_K \}$. It is required that the pairwise Wasserstein-2 distances should all be greater than 1.0. This is to make sure that observations of different clusters are statistically different.
    \item For each point $\posvec_{i,j}$, compute its Euclidean distance $d_{i,j}$ to the cluster center $\posvec_i$. For this point, generate a perturbed covariance matrix $P_{i,j}$ by adding Gaussian noise of scale $d_{i,j}\alpha$ to the ground-truth covariance matrix $\Sigma_i$. Sample a $D$-dimensional feature vector $\feavec_{i,j}$ from $P_{i,j}$. Similarly the maximum noise scale should be no larger than $10\%$ of the maximum entry in the ground-truth covariance matrix. Then the collection of all $(\feavec_{i,j}, \posvec_{i,j})$ makes the dataset.
\end{itemize}

Choice of \rebut{hyperparameters}: Larger $\alpha$ makes the synthetic data noisier and cluster boundaries fuzzier. Larger $k$ makes model estimation more accurate and stable, thus better clustering results.

\subsubsection{Real-world Datasets}


\textbf{(1) Pavement Dataset.} This dataset is a sensor-based, originally univariate time series collected by experts. Car sensors collect data while driving on different pavements (cobblestone, dirt and flexible). There are in total 1055 successive, variable-length subsequences of accelerometer readings sampled at 100 Hz. Each subsequence has a label from the aforementioned three pavement types. We use the first 10 entries of each subsequence as its feature vector, and treat the truncated data as a 1055-long, 10-dimensional multivariate time series. Our task is to put subsequences of the same pavement labels into the same clusters. 

The detailed information can be found at \url{https://timeseriesclassification.com/description.php?Dataset=AsphaltPavementType}. 

\textbf{(2) Vehicle Dataset.} This is a multivariate time series dataset collected by tracking the working status of commercial vehicles (specifically, dumpers) using smart phones and published in the literature. The original paper is here: http://kth.diva-portal.org/smash/record.jsf?pid=diva2

\textbf{(3) Gesture Dataset.} This dataset records hand-movements as multivariate time series. Each movement record is 315-time-step long, and each time-step has a 3-dimensional vector, representing the spatial coordinates of the center of the hand. There are in total 2238 records, each record 315-time-step long, thus the entire length of the dataset is 704,970 time-steps. All the records belong to one of the eight gestures. We randomly shuffle the order of the records, so that it is more challenging. Our task is to cluster time-steps into different gestures. The detailed information about this dataset can be found in \url{https://timeseriesclassification.com/description.php?Dataset=UWaveGestureLibrary}.

\textbf{(4) Climate Dataset.} This dataset consists of locations on the earth and their 5 climate attributes (temperature, precipitation, wind, etc.) based on the WorldClim database (\url{https://www.worldclim.org/data/worldclim21.html}). The ground-truth labels are the climate types of each location. There are in total 4741 locations, belonging to 14 different climate types. We use the great circle distance as the spatial distance metric for this dataset. 

\textbf{(5) iNaturalist-2018 Dataset.} This dataset contains images of species from all over the world together with their geotags (longitude and latitude). The entire dataset is huge and geospatially highly imbalanced (e.g., there are in total 24343 images in the test set, but 10792 out of them are in the contiguous US). We use the ImageNet-pretrained Inception V3 model to embed each image into a 2048-dimensional vector as \cite{mac2019presence,mai2023sphere2vec,mai2023csp} did, and reduce it to a 16-dimensional vector using PCA, for the sake of computability of STICC. The ground-truth labels of each image are hierarchical (i.e., from the top kingdom types to the bottom class types), and we use the 6 kingdom types as the cluster labels. Dataset (4) gives an example of spatially-constrained clustering in the multivariate raw feature space, and Dataset (5) extends the boundary to the latent representation space of images. 

\textbf{(6) (7) NYC Foursquare check-in dataset.} 
We use the NYC Check-in data proposed by \citep{6844862}. This dataset contains check-in data in New York City by Foursquare,
based on social media records. Each record includes VenueId, VenueCateg (POI Type), check-in timestamp (Weekday + Hour) and geospatial coordinate (Longitude + Latitude). We define the feature vector to be the normalized check-in vector, i.e., sum up the Hour attribute grouped by Week, and normalized this 7-dimensional vector. It is a feature vector representing the check-in patterns from Monday to Sunday. For evaluation, we construct 2 sets of ground-truth labels. 
One is from the \textbf{NYC Check-in data itself}: for each observation, we add up the one-hot POI type vectors of its nearest 50 neighbors and normalize it to be the POI embedding of this observation. Then, we cluster over these POI embeddings, and use the clustering labels as the ground-truth. Notice there is no information leak because our algorithm is fitted on check-in data and geo-coordinates only. 
The other is based on the \textbf{Primary Land Use Tax Lot Output (PLUTO) dataset} from NYC Open Data\footnote{\scriptsize{https://data.cityofnewyork.us/City-Government/Primary-Land-Use-Tax-Lot-Output-PLUTO-/64uk-42ks/data}}. We extract the land-use records and assign to each observation the nearest land-use record as its ground-truth land-use label. For the sake of data quality, we only use the records of Manhattan and Bronx.

\subsection{Further Discussion on Experiment Results}
\label{sec:more-exp-discussions}

\begin{figure*}[t!]
\centering
\subfloat[Climate dataset ground-truth clusters]{
\vspace{-0.4cm}
\includegraphics[width = 0.30\textwidth]{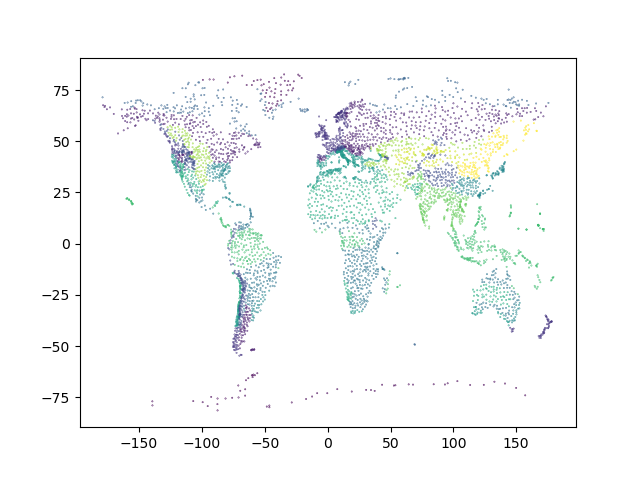}
\label{fig:climate-ground-truth}
}
\hspace{0.2cm}
\vspace{-0.2cm}
\subfloat[iNaturalist-2018 dataset ground-truth clusters (subset)]{
\vspace{-0.4cm}
\includegraphics[width = 0.30\textwidth]{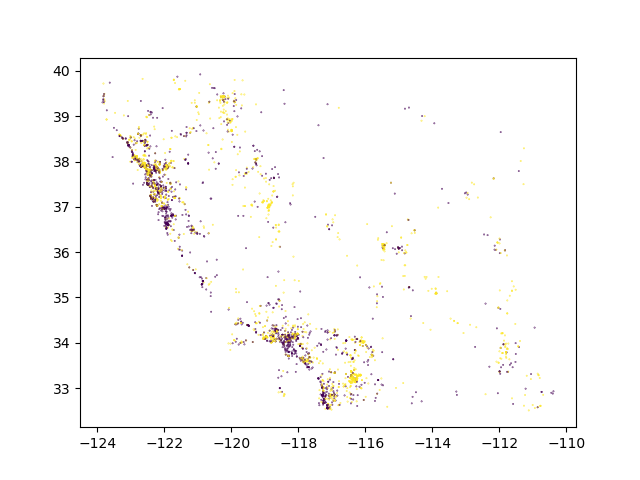}
\label{fig:inat-ground-truth}
}
\hspace{0.2cm}
\vspace{-0.2cm}
\subfloat[NYU POI/Land-use dataset]{
\vspace{-0.4cm}
\includegraphics[width = 0.30\textwidth]{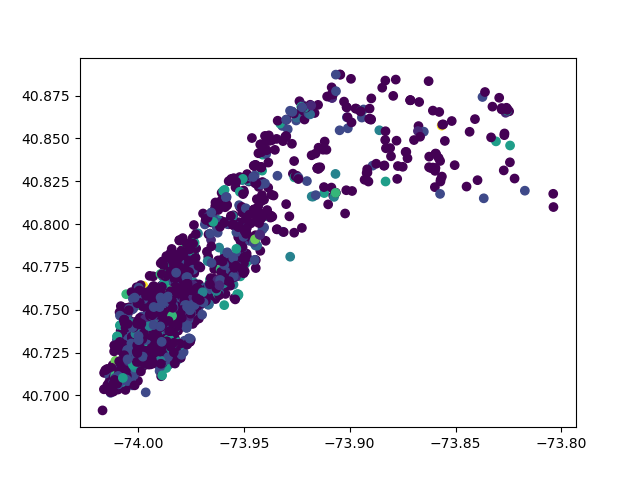}
\label{fig:poi-ground-truth}
}
\caption{Comparison of the Climate dataset, the iNaturalist-2018 dataset, the NYU POI/Land-use datasets. For visualization we only plot a subset of iNaturalist-2018 (California, two species). It is obvious that 1) the Climate dataset is very sparse and the ground-truth clusters have clear-cut borders, and 2) the iNaturalist/NYU datasets are dense and the ground-truth clusters overlap each other.}
\vspace{-0.4cm}
\label{fig:ground-truth-shape}
\end{figure*}


In Dataset (4) and Dataset (5) STICC/\loss{} without spatial constraints yield lower performance than GMM, because the spatial sampling rate is too low (i.e., there are too few data points within a unit distance).  
There is a dilemma to STICC/\loss{} algorithms: in order to obtain an adequate number of samples, we need to increase the sampling radius; however, as the sampling radius gets bigger, the samples become more noisy. In both cases the estimated distributions are inaccurate. 
Essentially, this problem originates from the balance between data sparsity (when having small neighborhood), and temporal/spatial incontiguity (when having large neighborhood)  
We address this problem by introducing a global prior. Since GMM can give a fairly good global estimation of the distribution of each cluster, we can use it as the prior distribution and update it in a maximum likelihood/Bayesian way given subsequence/subregion observations. This approach  demonstrates a large increase in clustering performance for iNaturalist-2018. Again it demonstrates how important spatial metric information is in clustering data points that satisfy local metric constraints. This finding may lead to future works.
\subsection{Theoretical Complexity and Empirical Runtime}
\label{sec:runtime}

\rebut{We denote $d$ as the data dimension, $n$ as the number of data points, and $K$ as the number of clusters. Theoretically, the complexity of \model{} is $O(n^2d^2)$. Firstly, we need to estimate covariance matrices for each data point, which is $O(n^2d\min(n,d))$. Since in most cases, $n >> d$, the complexity becomes $O(n^2d^2)$. After estimating the covariances, we compute the pairwise Wasserstein-2 distances, which is again $O(n^2d^2)$, because we need to do matrix multiplication ($O(d^2)$) $n^2$ times. Finally, we apply a distance-based clustering algorithm like DBSCAN on the Wasserstein-2 distance matrix, which is again $O(n^2)$. Thus the overall time complexity of \model{} is $O(n^2d^2)$. This means, theoretically the execution time of TICC/STICC is $C\cdot K$ times of that of \model{}.}

\rebut{Next, we show that the time complexity of the SOTA models (TICC and STICC) is $O(C\cdot K\cdot n^2d^2)$, where $C$ is how many iterations it takes to converge, which usually increases as $K$ and $n$ increase.}

\rebut{TICC/STICC needs to 1) compute an initial cluster assignment by kMeans, which is $O(n^2)$; 2) estimate cluster-wise covariance matrices and compute the likelihood of each data point against each cluster, which is $O(K\cdot n^2d^2)$; 3) update cluster assignment, which is reported $O(K\cdot n)$ in the original papers; 4) repeat (1) to (3) $C$ times until convergence. Thus the overall time complexity is $O(C\cdot K\cdot n^2d^2)$.}

\rebut{We also evaluated the empirical time complexity of each clustering algorithm. Please refer to the “RT” column in Table \ref{tab:runtime}. We can see that TICC/STICC is much slower than our \model{}. Notice the time TICC/STICC takes highly depends on how many iterations it takes to converge.}

\rebut{Finally, the spatial complexity of both \model{} and TICC/STICC is $O(n\cdot d^2)$, since all we need to store is the covariance matrices of each data point.}

\begin{table*}[t!]
  \caption{
  Performance and runtime comparison across different model-based clustering algorithms on 1-D (temporal) and 2-D (spatial) real-world datasets. 
  $d$ denotes the feature dimension, $c$ denotes the ground-truth cluster number, and $N$ denotes the size of each dataset. RT denotes the average run-time in seconds.
  \textbf{Bold} numbers and \rebut{\underline{underlined}} numbers indicate the best and second best performances. 
  TICC applies to 1-D datasets and STICC applies to 2-D datasets. $\beta_0$ means there is no temporal/spatial penalty term applied. NC means the algorithm does not converge.
  \loss-wo/\loss-w represents \loss{} loss without/with metric information respectively. 
  }
  \label{tab:runtime}
  \centering \tiny 
  \setlength{\tabcolsep}{3pt}
  {\small 
  \begin{tabular}{*{16}{c}}
    \toprule
    \midrule
    \multirow{5}{*}{Model} &  \multicolumn{9}{c}{Temporal Dataset (1-D)} &  \multicolumn{6}{c}{Spatial Dataset (2-D)} \\
    \cmidrule{2-16}
     & \multicolumn{3}{c}{\begin{tabular}{@{}c@{}}Pavement \\ \begin{tabular}{@{}c@{}}$d$=10, $c$=3 \\ N=1,055 \end{tabular} \end{tabular}} & \multicolumn{3}{c}{\begin{tabular}{@{}c@{}}Vehicle \\ \begin{tabular}{@{}c@{}}$d$=7, $c$=5 \\ N=16,641 \end{tabular} \end{tabular}} &\multicolumn{3}{c}{\begin{tabular}{@{}c@{}}Gesture \\ \begin{tabular}{@{}c@{}}$d$=3, $c$=8 \\ N=704,970 \end{tabular} \end{tabular}} & \multicolumn{3}{c}{\begin{tabular}{@{}c@{}}Climate \\ \begin{tabular}{@{}c@{}}$d$=5, $c$=14 \\ N=4,741 \end{tabular} \end{tabular}} & \multicolumn{3}{c}{\begin{tabular}{@{}c@{}}iNat2018 \\ \begin{tabular}{@{}c@{}}$d$=16, $c$=6 \\ N=24,343 \end{tabular} \end{tabular}}\\
    \cmidrule{2-16}
     & ARI & NMI & RT & ARI & NMI & RT & ARI & NMI & RT & ARI & NMI & RT & ARI & NMI & RT\\
    \midrule
    GMM & 28.05 & 28.74 & < 1s & 57.87 & 58.78 & 3s & 2.44 & 4.15 & 14s & \underline{19.06} & 34.97 & < 1s & 21.72 & 35.91 & 9s \\
    (S)TICC-$\beta_0$ & 58.54 & 58.83 & 383s & 40.12 & 45.86 & 441s & 3.26 & 6.56 & 4782s & 13.30 & 30.53 & 1277s & NC & NC & 6881s \\
    (S)TICC & 62.27 & 61.89 & 508s & 50.53 & 53.68 & 566s & \underline{12.20} & 23.20 & 4511s & 17.62 & \underline{37.29} & 1204s  & NC & NC & 6325s \\
    \midrule
    \loss-wo & \underline{76.10} & \underline{74.36} & 14s & \underline{63.31} & \underline{58.60} & 74s & 8.12 & \underline{33.60} & 573s & 16.63 & 36.73 & 746s & \underline{21.90} & \underline{36.47} & 588s \\
    \loss-w & \textbf{77.64} & \textbf{77.22} & 14s & \textbf{65.04} & \textbf{59.36} & 76s & \textbf{26.51} & \textbf{55.34} & 554s & \textbf{20.08} & \textbf{40.91} & 755s & \textbf{42.70} & \textbf{40.49} & 594s \\
    \hline
    \hline
  \end{tabular}
  }
\vspace{-0.3cm}
\end{table*}
\subsection{Hyperparameter Tuning}
\label{sec:tune}

\begin{figure*}[ht!]
\centering
\vspace{-0.2cm}
\subfloat[\rebut{Tuning the number of neighbors $n$ }]{\includegraphics[width = 0.3\textwidth]{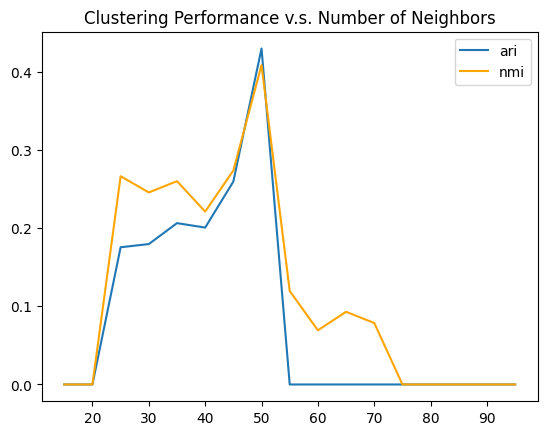}\label{fig:n-tuning}}
\subfloat[\rebut{Tuning the weight $\beta$ }]{\includegraphics[width = 0.3\textwidth]{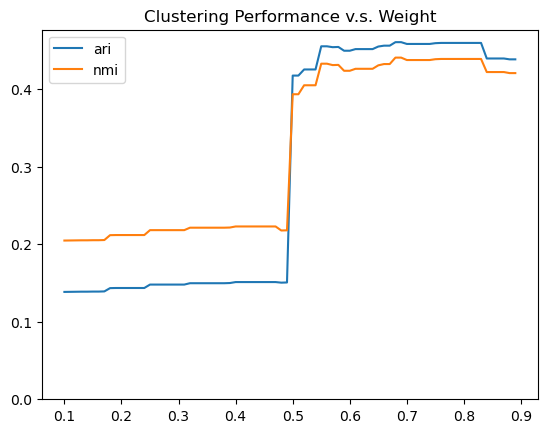}\label{fig:beta-tuning}}
\subfloat[\rebut{Tuning the margin $\delta$ }]{\includegraphics[width = 0.3\textwidth]{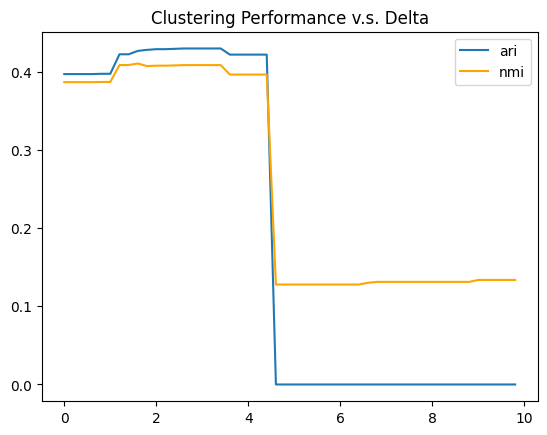}\label{fig:delta-tuning}}
\vspace{-0.2cm}
\caption{\rebut{The performance curve with regard to the grid-searched hyperparameters $n$, $\beta$ and $\delta$}}
\label{fig:hyperparameters}
\end{figure*}

\rebut{We include an ablation study to investigate the influence of the number of neighbors $n$, the weight $\beta$, and the margin $\delta$ using the most complicated iNaturalist 2018 dataset. Figure \ref{fig:hyperparameters} demonstrates that the search space of single hyperparameters has good convexity. Thus, we can easily and quickly tune the hyperparameters by hierarchical grid search.}

\rebut{For tuning $\beta$ and $\delta$, we do not need to re-compute the covariance matrices. Instead, we only need to re-run the density-based clustering algorithm, such as DBSCAN. Thus the time complexity of a complete grid search is only $O(A\cdot B \cdot C n^2)$, where A and B are the grid sizes of $n$, $\beta$ and $\delta$.} 

\rebut{Unlike \model{}, the competing baselines TICC/STICC must re-run the entire algorithm when tuning hyperparameters. That means the complete grid search is $O(A\cdot B\cdot C\cdot K\cdot n^2d^2)$, even if we only tune the most important $\lambda$ and $\beta$ hyperparameters.
}

\begin{figure*}[ht!]
\centering
\subfloat[{The average value of the first  and  second term in Eq~\ref{eq:mcmb-as-test-simp}}]{
\vspace{-0.2cm}
\includegraphics[width = 0.45\textwidth]{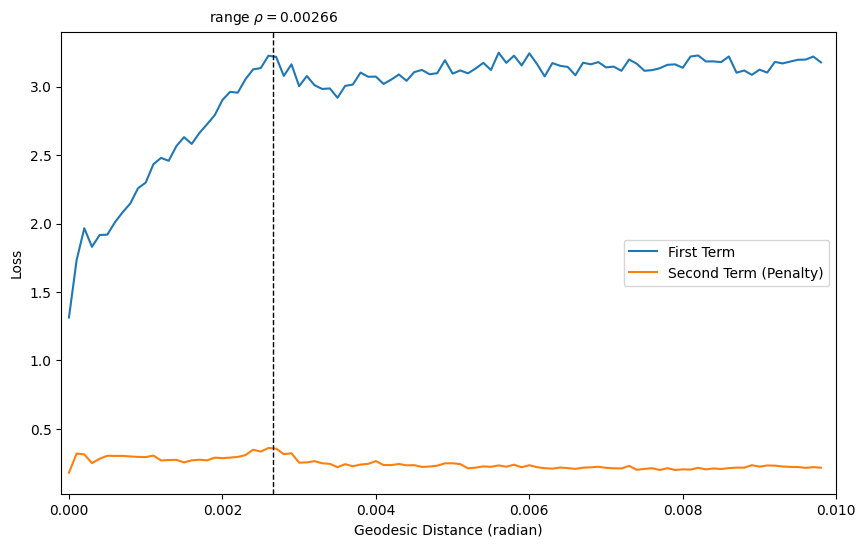}
\label{fig:loss-analysis}
}
\hspace{0.4cm}
\vspace{-0.2cm}
\subfloat[
{The contribution of the second term in Eq~\ref{eq:mcmb-as-test-simp} to the total loss.}]{
\vspace{-0.4cm}
\includegraphics[width = 0.45\textwidth]{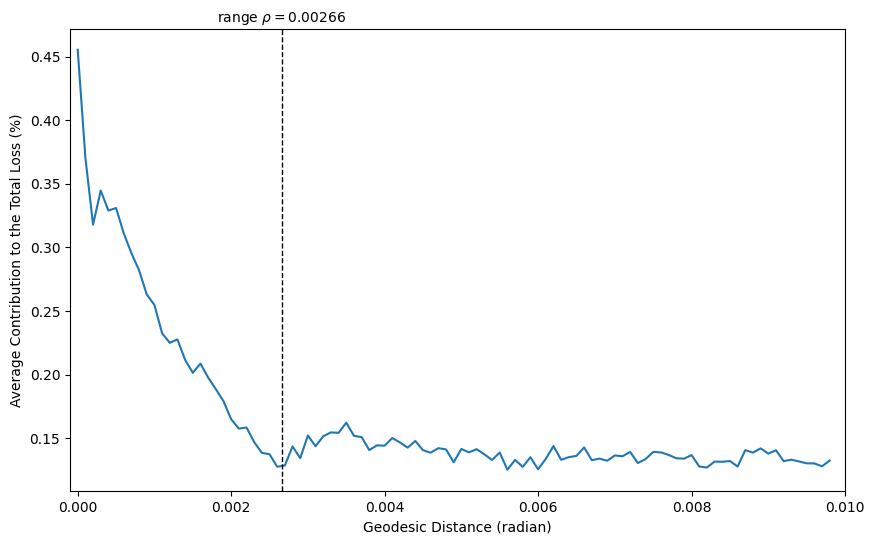}
\label{fig:range-analysis}
}
\caption{{Analysis of the loss composition in Eq~\ref{eq:mcmb-as-test-simp}. The average contribution of the metric-constraint penalty term to the total loss beyond the range quickly drops down to below $15\%$, which can be ignored in practice. 
}}
\vspace{-0.5cm}
\label{fig:range-condition}
\end{figure*}



\end{document}